\documentclass{article} % For LaTeX2e
\usepackage{iclr2024_conference, times}

\usepackage{amsmath,amsfonts,bm}

\def\eqref#1{equation~\ref{#1}}
\def\1{\bm{1}}

\DeclareMathAlphabet{\mathsfit}{\encodingdefault}{\sfdefault}{m}{sl}
\SetMathAlphabet{\mathsfit}{bold}{\encodingdefault}{\sfdefault}{bx}{n}

\usepackage{hyperref}
\usepackage{url}

\usepackage[utf8]{inputenc} % allow utf-8 input
\usepackage[T1]{fontenc}    % use 8-bit T1 fonts
\usepackage{hyperref}       % hyperlinks
\usepackage{url}            % simple URL typesetting
\usepackage{booktabs}       % professional-quality tables
\usepackage{amsfonts}       % blackboard math symbols
\usepackage{nicefrac}       % compact symbols for 1/2, etc.
\usepackage{microtype}      % microtypography
\usepackage{xcolor}         % colors
\usepackage{bbm}
\usepackage{graphicx}
\usepackage{caption}
\usepackage{subcaption}
\usepackage{wrapfig}
\usepackage{multirow}
\usepackage{comment}
\usepackage{enumitem}

\newtheorem{definition}{Definition}
\newtheorem{lemma}{Lemma}

\newtheorem{Proposition}{Proposition}

\RequirePackage{algorithm}
\RequirePackage{algorithmic}
\usepackage{amsmath}
\usepackage{enumitem,amssymb}
\newlist{todolist}{itemize}{2}
\setlist[todolist]{label=$\square$}
\PassOptionsToPackage{numbers, compress}{natbib}

\author{Aniket Roy, Maiterya Suin, Anshul Shah, Ketul Shah, Jiang Liu, Rama Chellappa \\
Johns Hopkins University\\
}

\iclrfinalcopy % Uncomment for camera-ready version, but NOT for submission.
\title{DiffNat: Improving diffusion image quality using natural image statistics}

\begin{document}

\maketitle

\begin{abstract}

Diffusion models have advanced generative AI significantly in terms of editing and creating naturalistic images.
\textcolor{blue}{However, efficiently improving generated image quality is still of paramount interest.}
In this context, we propose a generic ``naturalness'' preserving loss function, viz., kurtosis concentration (KC) loss, which can be readily applied to any standard diffusion model pipeline to elevate the image quality. 
Our motivation stems from the projected kurtosis concentration property of natural images, which states that natural images have nearly constant kurtosis values across different band-pass versions of the image. To retain the ``naturalness'' of the generated images, we enforce reducing the gap between the highest and lowest kurtosis values across the band-pass versions (e.g., Discrete Wavelet Transform (DWT)) of images. Note that our approach does not require any additional guidance like classifier or classifier-free guidance to improve the image quality. We validate the proposed approach for three diverse tasks, viz., (1) personalized few-shot finetuning using text guidance, (2) unconditional 
image generation, and (3) image super-resolution. Integrating the proposed KC loss has improved the perceptual quality across all these tasks in terms of both FID, MUSIQ score, and user evaluation. 

\end{abstract}

%%%%%%%%% BODY TEXT

\section{Introduction}

\begin{figure}
    \centering
    \includegraphics[scale=0.75]{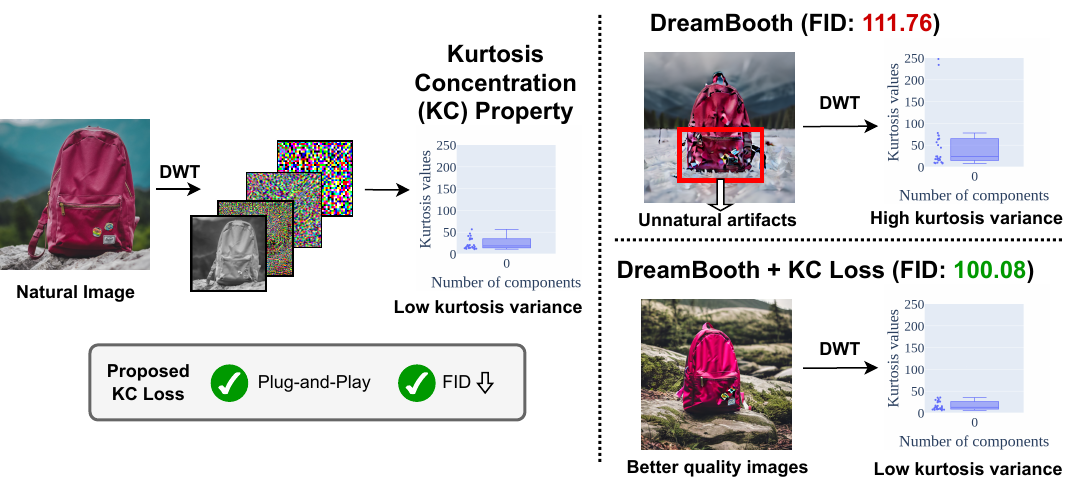}
    \vspace{-0.2cm}
    \caption{\small{Overview of DiffNat. We utilize the kurtosis concentration (KC) property of natural images, which states the kurtosis values across different bandpass filtered (Discrete Wavelet Transform (DWT)) version of the images tend to be constant (left).
    As can be observed in this figure, 50 percentile of the kurtosis values reside in the blue box, which indicates the concentration of the kurtosis values. 
    For natural images, this spread is relatively smaller.
    Inspired by this observation, we propose a novel KC loss, which minimize the deviation of kurtosis across different bandpass (DWT) versions of images. 
    This loss can be applied to any diffusion pipleline with the traditional reconstruction loss.
    \textcolor{blue}{Here, we consider DreamBooth~\cite{ruiz2022dreambooth}. DreamBooth generated images might have unnatural artifacts, producing high kurtosis variance (large spread in the blue box) and higher FID (right top).} The FID over the DreamBooth dataset is reported here. Adding KC loss improves image quality in terms of FID and reduces the kurtosis variance (right bottom).} } 
\label{fig:diffnat_teaser}
\vspace{-0.65cm}
\end{figure}

Multi-modal generative AI has advanced by leaps and bounds with the advent of the diffusion model. Large-scale text-to-image diffusion models, e.g., DALLE~\cite{ramesh2022hierarchical}, Stable-diffusion~\cite{rombach2022high} etc. synthesize high-quality images in diverse scenes, views, and lighting conditions from text prompts. The quality and diversity of these generated images are astonishing since they have been trained on a large collection of image-text pairs and are able to capture the visual-semantic correspondence effectively.
Although the diffusion model-generated images look realistic, a recent study has shown that the generated images can be distinguished from natural images using state-of-the-art image forensic tools~\cite{corvi2023detection}. This implies that state-of-the-art generative models might be good at image editing, but often leave unnatural traces and lack ``naturalness'' quality. 
This problem is more prevalent in the cases of few-shot finetuning of large multi-modal diffusion models, e.g., ``personalization'' of text-to-image diffusion model. 
Popular methods, e.g., DreamBooth~\cite{ruiz2022dreambooth}, Custom diffusion~\cite{kumari2022multi}, etc. achieve impressive subject-driven ``personalized'' image generation based on text prompts, but these have several limitations, e.g., image quality degradation due to unnatural artifacts, etc. 
Image quality is of utmost importance for other generative tasks as well, e.g., super-resolution, image restoration, unconditional image generation, etc. Some examples of unnatural artifacts are shown in Fig.~\ref{fig:diffnat_teaser}. 

% In this task, subject specific identifiers have to be learnt from few-examples by finetuning the text-to-image diffusion model in order to generate variations with respect to pose, background and color with that particular subject from corresponding text-prompts. 
% The primary challenge is to preserve subject fidelity in the visual space, while background is customized based on the text-prompt in diverse conditions.

To improve image quality, several methods rely on guidance methods, e.g., classifier guidance, and classifier-free guidance~\cite{dhariwal2021diffusion} etc. However, these methods require external supervision and add complexity to the training process.
Our goal is to improve the image quality without any additional guidance, yet preserving the ``naturalness'' of the generated images by exploring the well-known kurtosis concentration property of natural images~\cite{zhang2014using}.  
This property states that natural images have nearly constant kurtosis (fourth order moment) values across different band-pass (e.g., Discrete Cosine Transform (DCT), Discrete Wavelet Transform (DWT)) versions of the images ~\cite{zhang2014using}. 
Inspired by this property, we propose a novel kurtosis concentration (KC) loss, which is generic and applicable to any diffusion based pipeline. More specifically, this loss minimizes the gap in the kurtosis of an image across band-pass filtered versions and thus enforce the ``naturalness'' of the generated images.

% We have integrated this loss in different generative tasks, e.g, 
This loss is general-purpose and does not even require any labels. It can be adapted to various generative tasks with minimal effort. In this work, we experiment with diverse tasks of:
(1) personalized few-shot finetuning of text-to-image diffusion model, (2) unconditional image generation, and (3) image super-resolution.  

Our major contributions are as follows:
\begin{itemize}[noitemsep, topsep=0pt,leftmargin=*]
    \item We introduce DiffNat - a framework for improving the image quality of diffusion models using natural image statistics.
    Based on the kurtosis concentration property, we propose a novel loss function by minimizing the gap of kurtosis values (i.e., the difference between maximum and minimum kurtosis values) across the band-pass (in DWT domain) filtered version of the image. To the best of our knowledge, we are the first to propose this loss based on natural image statistics.
    \item We provide theoretical insights into the proposed loss function for generating images with better perceptual quality. 
    \item We validate the proposed loss in diverse generative tasks, e.g., (1) personalized few-shot finetuning of text-to-image diffusion model using text guidance, (2) unconditional image generation, and (3) image super-resolution. Experiments suggest that incorporating the proposed loss improves the perceptual quality in all these tasks across different benchmarks. We have validated the proposed approach with a user study as well. 
\end{itemize}
 
\section{Related Work}

\noindent\textbf{Deep Generative Models.} 
Generative models (GANs \cite{goodfellow2020generative}, VAEs \cite{kingma2019introduction}, flow-based models \cite{rezende2015variational}, and diffusion models \cite{ho2020denoising}) learn the probability distribution of given data, allowing us to sample new data points from the distribution. Deep generative models have been used for modeling the distribution of faces \cite{karras2019style}, 3D objects \cite{wu2016learning}, videos by \cite{vondrick2016generating}, natural images by \cite{karras2019style, brock2018large}, etc for unconditional synthesis. Conditioning the generative models on segmentation mask \cite{isola2017image}, class label \cite{mirza2014conditional}, text \cite{tao2022df} enables us to have more control over the generated images. 
Generative models can be controlled using guidance from images, texts, etc. ILVR \cite{choi2021ilvr} present an iterative way to guide the image synthesis process using a reference image. Instance-conditioned GAN \cite{casanova2021instance} allows for generating semantic variations of a given reference image, by training using the nearest neighbors of the reference image. \cite{roich2022pivotal} fine-tune the generator around an inverted latent code anchor, allowing for latent-based semantic editing on images that are out of the generator’s domain. 
% \cite{nitzan2022mystyle} creates a personalized generative prior using a few images of a subject, which allows reconstructing and editing of the given subject. These methods are still limited to structured domains such as facial images.  

\noindent\textbf{Text-to-Image Generation and Editing.}
Generating high fidelity, diverse images using text inputs has seen tremendous progress recently. Many approaches based on GANs have been proposed for text-to-image generation \cite{qiao2019mirrorgan, tao2022df, liao2022text, zhu2019dm, ruan2021dae}. More recent advances in text-based image synthesis (Stable Diffusion \cite{rombach2022high}, Imagen~\cite{saharia2022photorealistic}, etc) have been powered by diffusion models trained on massive datasets.
%trained on large-scale image-text data have also unlocked many applications in image manipulation and editing, apart from remarkable semantic generation. 
% Even though these methods can produce high quality images from text and/or image prompts, editing images with fine-grained control still remains challenging.  
GAN-based text-based image editing approaches \cite{crowson2022vqgan, bau2021paint, abdal2022clip2stylegan, gal2021stylegan, patashnik2021styleclip} have made significant strides recently thanks to CLIP \cite{radford2021learning}. Diffusion-based text-to-image editing methods \cite{ruiz2022dreambooth, kumari2022multi, gal2022image} show better control and impressive editing results. 
% Recently, there have been efforts for such few-shot “personalization” of these powerful text-to-image models.
% “Personalizing” these text-to-image models using few examples still remains challenging because the resulting image needs to preserve the subject involved, while the resulting image has to also appear natural. 
For “personalizing” these text-to-image models, Textual Inversion \cite{gal2022image} represent a subject as a new "word" in the embedding space of a diffusion model, which is used in natural language prompts for creating new images of the subject in novel scenes. DreamBooth \cite{ruiz2022dreambooth} embeds the subject in the output domain of the model and the resulting unique identifier is used to synthesize novel images of the subject in unseen contexts. Custom Diffusion \cite{kumari2022multi} extends this by enabling the composition of multiple new concepts with existing ones. 
% Though these methods are able to preserve the identity of the subject, they do not enforce any constraints for the synthetic image to be natural, which results in unnatural artefacts in the generated images. 
%Our method includes a novel loss for preserving the ``naturalness" of the images to mitigate these issues. 

% \begin{itemize}
%     \item DreamBooth
%     \begin{itemize}
%         \item “personalization” of text-to-image diffusion models.
% subject is embedded in the output domain of the model, the unique identifier can be used to synthesize novel photorealistic images of the subject contextualized in different scenes.
%         \item in diverse scenes, poses, views and lighting conditions that do not appear in the reference images. 
%     \end{itemize}
%     \item Custom Diffusion 
%     \begin{itemize}
%         \item compositional fine-tuning of multiple concepts
%         \item we only fine-tune a subset of cross-attention layer parameters
%     \end{itemize}
%     \item Textual Inversion
%     \begin{itemize}
%         \item Introducing and optimizing a word vector for each new concept
%     \end{itemize}
%     \item An image is worth one word 
%     \begin{itemize}
%         \item a method to represent visual concepts, like an object or a style, through new tokens in the embedding space of a frozen text-to-image model, resulting in small personalized token embeddings
%     \end{itemize}
% \end{itemize}

\noindent\textbf{Natural Image Statistics.} Natural images have interesting scale-invariance and noise properties~\cite{zoran2009scale}, which has been used for image restoration problems. 
Projected kurtosis concentration property of natural images, i.e., natural images tend to have constant kurtosis values across different band-pass (DCT, DWT) filtered version has been used for blind forgery detection~\cite{zhang2014using}. 
% Inspried by these observation, we analyse the noise variance properties of natural and diffusion model generated images and propose a novel loss function based on natural image statistics for generating better quality images.
  
%\input{motivation}
\section{Method}

\begin{wrapfigure}{r}{0.35\textwidth}
    \centering
    \vspace{-10pt}
    \includegraphics[width=0.4\textwidth]{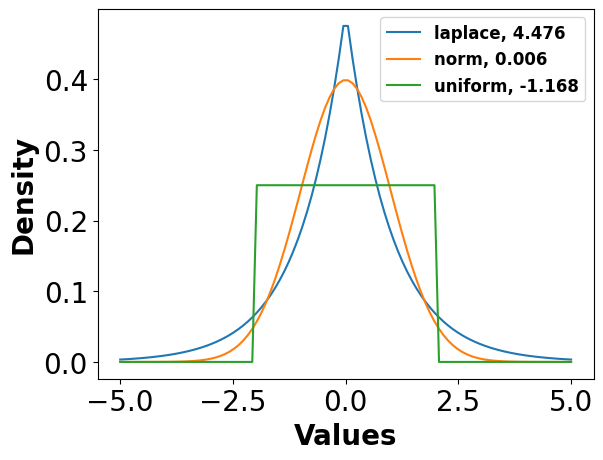}
    % \vspace{-10pt}
    \caption{\small{Kurtosis of various distributions. Intuitively, kurtosis captures the peakedness of the distribution.}} 
    \label{fig:kurtosis}
    \vspace{-20pt}
\end{wrapfigure}

In this section, we present the concept of the kurtosis concentration loss, which can be applied to various generative tasks for enhancing the quality of generated images.
We start with providing a basic understanding of the kurtosis concentration property of natural images and how we leverage this property to propose kurtosis concentration (KC) loss which enforces the ``naturalness'' of the generated samples.

\subsection{Kurtosis Concentration Property}

% \begin{figure}
%     \centering
%     \includegraphics[scale=0.4]{figures/kurtosis_dist.png}
%     \caption{Kurtosis of distributions.} 
%     \label{fig:kurtosis}
% \end{figure}

%\subsection{Kutosis Concentration Property}

% \begin{minipage}{\textwidth}
% \begin{minipage}[b]{0.49\textwidth}
% \centering
% \includegraphics[scale=0.3]{figures/kurtosis_dist.png}
% % \caption{Kurtosis of distributions.} 
% \label{fig:kurtosis}
% \end{minipage}

% \begin{minipage}[b]{0.49\textwidth}
% \centering
% \begin{tabular}{lccc}
%   \toprule
%   {\bf Method} & {\bf MUSIQ score $\uparrow$} & {\bf Noise $\sigma^2$ $\downarrow$} \\ 
%   \midrule
%   {DreamBooth~\cite{ruiz2022dreambooth}} & {68.319} & {0.382} \\
%   {DiffNat} & {68.924} & {0.325} \\
%   {Natural images} & {\bf 72.150} & {\bf {3.24 exp(-47)}} \\  	  
%   \bottomrule
%   \end{tabular}
% \end{minipage}
% \end{minipage}

\begin{definition}
\label{def:kurtosis}
Kurtosis : Kurtosis is a measure of the ``peakedness'' of the probability distribution of a random variable~\cite{zhang2014using}.
For a random variable $x$, its kurtosis is defined as,
\begin{equation}
    \kappa(x) = \frac{\mu_4{(x)}}{(\sigma^2(x))^2} - 3.
\end{equation}
where $\sigma^2(x) = \mathbb{E}_{x}[ (x - \mathbb{E}_{x}(x))^2]$ and $\mu_{4}(x) = \mathbb{E}_{x}[ (x - \mathbb{E}_{x}(x))^4]$ is the second order and fourth order moment of $x$. E.g., Gaussian random variable has kurtosis value 0. 
\end{definition}

Intuitively, kurtosis is a measure of the peakedness of a distribution. Kurtosis of well-known distributions is shown in Fig.~\ref{fig:kurtosis}. A positive kurtosis indicates that the distribution is more peaked than the normal distribution and negative kurtosis indicates it to be less peaked than normal distribution~\cite{zhang2014using}. Kurtosis is a useful statistic used for blind source separation~\cite{naik2014blind} and independent component analysis (ICA)~\cite{stone2002independent}.

For a random vector $x$, we define the kurtosis of the 1D projection of $x$ onto a unit vector $w$ as  projection kurtosis, i.e., $\kappa(w^Tx)$.
This projection kurtosis is an effective measure for the statistical properties of high-dimensional data.
E.g., if $x$ is a Gaussian, its projection over any $w$ has a 1D Gaussian distribution. Therefore, its projection kurtosis is always zero, which exhibits the kurtosis concentration (to a single value, i.e., zero) of Gaussian.

% %%%%%%%%%%%%%%%%% earlier %%%%%%%%%%%%%%%%%%%%%%%%%
% \begin{lemma}
% \label{lemma: lemma1}
%   If the noisy version of the natural image is denoted by y = x + n, where x is the natural image and n is a white Gaussian noise with zero mean and variance $\sigma^2$, then the kurtosis of y, $\kappa(y)$ can be expressed as,
% \begin{equation}
%     \kappa(y) = \kappa(x) (1 - \frac{1}{\text{SNR}(y)})^2 
% \end{equation}

% where Signal-to-Noise Ratio (SNR) is defined as, $SNR(y) = \frac{\sigma^2(y)}{\sigma^2(n)}$. 
% \end{lemma}

% \textit{Proof.} The proof is provided in the supplementary material.

% \begin{Proposition}
% \label{Proposition: theorem1}
% Minimizing kurtosis improves image quality.
% \end{Proposition}

% % \textit{Proof.}
% From Lemma 1, we can observe there exists an inverse relation between the kurtosis and image quality (SNR), which implies minimizing kurtosis improves image quality.
% % Also, from this equation, we conclude that minimizing kurtosis minimizes the noise variance.

% %%%%%%%%%%%%%%%%%%%%%%%%%%%%%%%%%%%%%%%%%%%%%%%%%%%%%

It is well-known that natural images can be modeled using zero-mean GSM vector~\cite{zoran2009scale}. Next, we analyze an interesting property of the GSM vector. 

%%%%%%%% added details %%%%%%%%%%%%%%

\begin{lemma}
\label{lemma: lemma2}
A Gaussian scale mixture (GSM) vector $x$ with zero mean has the following probability density function:
\begin{equation}
    p(x) = \int_{0}^{\infty} \mathcal{N}(x;0,z\Sigma_{x})p_{z}(z)dz
\end{equation}
and its projection kurtosis is \underline{constant} with respect to the projection direction w, i.e.,
\begin{equation}
    \kappa(w^Tx) = \frac{3var_{z}\{z\}}{\mathcal{E}_{z}\{z\}^2}
\end{equation}
where $\mathcal{E}_{z}\{z\}$ and $var_{z}\{z\}$ are the mean and variance of latent variable $z$ respectively.
\end{lemma}

\textit{Proof.} The proof is provided in the supplementary material.

This result by~\cite{zhang2014using} shows that projection kurtosis is constant across projection directions (e.g., wavelet basis), which provides a theoretical understanding of the kurtosis concentration property, which we will discuss next.

\textbf{Kurtosis Concentration Property}: It has been observed that for natural images, kurtosis values across different band-pass filter channels tend to be close to a constant value. This is termed as kurtosis concentration property of natural images~\cite{zhang2014using, zoran2009scale}. 
It can be interpreted as an implication of Lemma 1, if we consider natural images as zero-mean GSM vector. 
As a motivating example, we demonstrate the kurtosis concentration property of natural images in Fig.~\ref{fig:diffnat_teaser}. Next, we establish the relation between the projection kurtosis of the noisy version of the image and the corresponding signal-to-noise ratio.

\begin{lemma}
\label{lemma: lemma3}
  If the noisy version of the natural image is denoted by, y = x + n, where x is a whitened GSM vector (normalized natural image) and n is a zero-mean white Gaussian noise with variance $\sigma^2I$, x and n are mutually independent of each other, then the projection kurtosis of y, $\kappa(w^Ty)$ can be expressed as:
\begin{equation}
    \kappa(w^Ty) = \kappa(w^Tx) \Big(1 - \frac{c}{SNR(y)} \Big)^2 = \frac{3var_{z}\{z\}}{\mathcal{E}_{z}\{z\}^2} \Big(1 - \frac{c}{SNR(y)} \Big)^2
\end{equation}

where Signal-to-Noise Ratio (SNR) is defined as, $SNR(y) = \frac{\sigma^2(y)}{\sigma^2(n)}$ and $c$ is a constant.
\end{lemma}

\begin{wrapfigure}{r}{0.5\textwidth}
    \centering
    \includegraphics[width=0.5\textwidth]{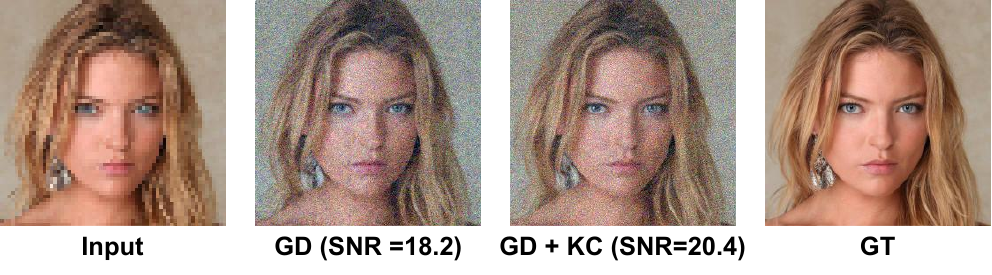}
    \vspace{-10pt}
    \caption{\small{Empirical evidence of proposition 1, i.e., minimizing KC loss denoise input signal. Here we take standard guided diffusion (GD) model with and without kurtosis loss for inference for 400 steps, and the denoised outputs are shown. The model trained with KC loss generates better quality images, which is also reflected in higher SNR values. GT refers to ground truth.}} 
    \label{fig:prop_1_evidence}
    \vspace{-20pt}
\end{wrapfigure}

\textit{Proof.} The proof is provided in the supplementary material.

This result utilizes the fact that, natural images have constant projection kurtosis, stated in Lemma 1. Next, we connect projection kurtosis minimization to denoising.

\begin{Proposition}
\label{Proposition: theorem1}
Minimizing projection kurtosis denoise input signal.
\end{Proposition}

From Lemma 2, we can observe there exists an inverse relation between the projection kurtosis and image quality (SNR), therefore minimizing projection kurtosis will increase SNR and the image will be denoised better. 

The primary objective of diffusion models is to learn denoising from a noisy image or latent embedding in order to generate a clean image. 
Then by Lemma 2, the projection kurtosis minimization results in better denoising (high SNR) of the reconstructed image.
In the case of diffusion models, the underlying denoising UNet is trained using mean squared error objective w.r.t the reconstructed image and the clean image. During inference, the reconstructed image is iteratively denoised and refined for $T$ steps to generate the final image with higher quality. Therefore, adding an objective to minimize the projection kurtosis of the reconstructed image, i.e., increasing the SNR (Lemma 2) would effectively lead to better denoising at each step and the final image would be of improved quality (shown in Fig.~\ref{fig:prop_1_evidence}).

\begin{figure}
    \centering
    \includegraphics[scale=0.65]{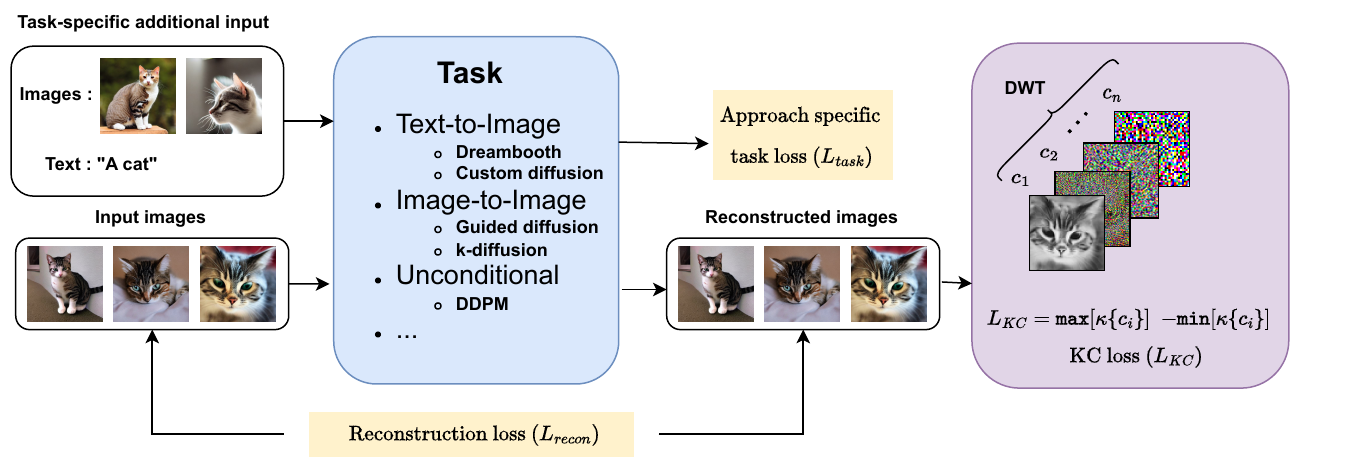}
    \caption{\small{Overview of DiffNat. The proposed kurtosis concentration loss can be integrated to any diffusion based approach for various tasks (e.g., text-to-image generation (DreamBooth, Custom diffusion), super-resolution image-to-image generation (Guided diffusion, k-diffusion), unconditional image generation (DDPM)). In addition to the task specific losses, and general reconstruction loss, we incorporate the kurtosis concentration loss ($L_{KC}$), which operates on the reconstructed images and minimize the kurtosis deviation (i.e., max[$\kappa[\{c_i\}$] - min[$\kappa[\{c_i\}$]) across Discrete Wavelet Transform (DWT) filtered version of the reconstructed image, Here, $c_1$, $c_2$ .. are DWT filtered version of the reconstructed image and $\kappa(x)$ denote kurtosis of $x$.}} 
    \vspace{-0.2cm}
    \label{fig:overview}
\end{figure}

\subsection{Kurtosis Concentration (KC) loss}

% Here, we introduce the proposed kurtosis-concentration loss, which enforces the generated images to be more ``natural'' looking based on properties of natural image statistics. In sec. [], we discussed the kurtosis concentration property of natural images, according to which the kurtosis values of natural images across In different bandpass filters tend to be a constant value. 
In this work, we leverage this property to introduce a novel loss function, viz., Kurtosis Concentration loss (KC loss) for training deep generative models.
Unlike prior approaches~\cite{zhang2014using}, where the KC property has been used for noise estimation, source separation, etc., we utilize this property of natural images as a prior to train generative models for generating images with better perceptual quality. To validate our loss, we experiment with state-of-the-art generative models, i.e., diffusion models. Note that, our proposed loss can be integrated as a plug-and-play with any diffusion pipeline. We describe the basic diffusion pipeline and KC loss as follows.

\begin{figure}
    \centering
    \begin{subfigure}{0.18\textwidth}
       \includegraphics[width=\linewidth]{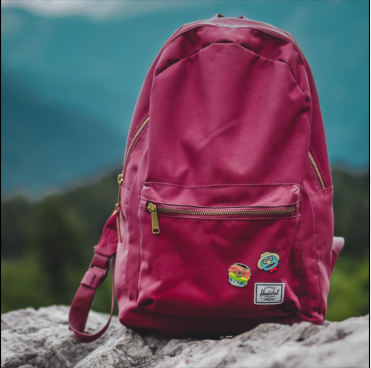}
        \caption{Natural image}
    \end{subfigure}
    \begin{subfigure}{0.18\textwidth}
        \includegraphics[width=\linewidth]{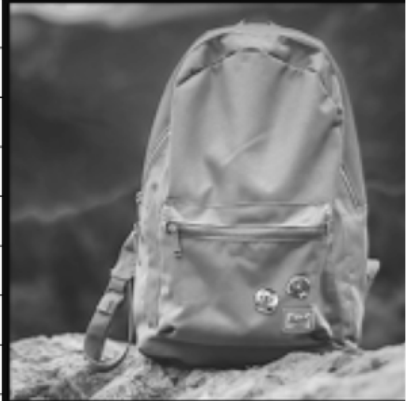}
        \caption{LL subband}
    \end{subfigure}
    \begin{subfigure}{0.18\textwidth}
        \includegraphics[width=\linewidth]{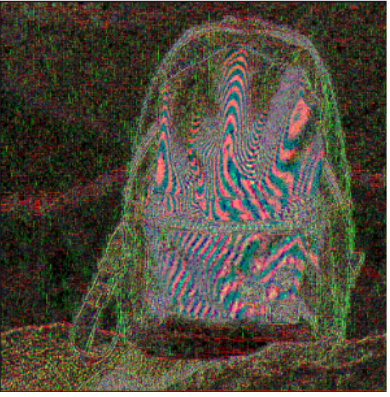}
        \caption{LH subband}
    \end{subfigure}
    \begin{subfigure}{0.18\textwidth}
        \includegraphics[width=\linewidth]{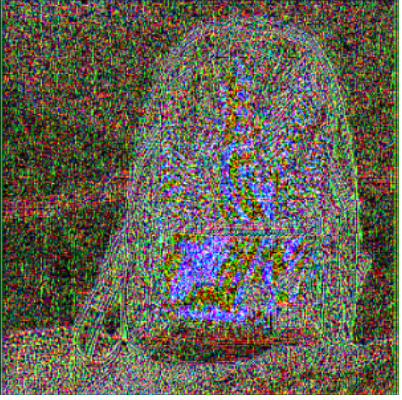}
        \caption{HL subband}
    \end{subfigure}
    \begin{subfigure}{0.18\textwidth}
       \includegraphics[width=\linewidth]{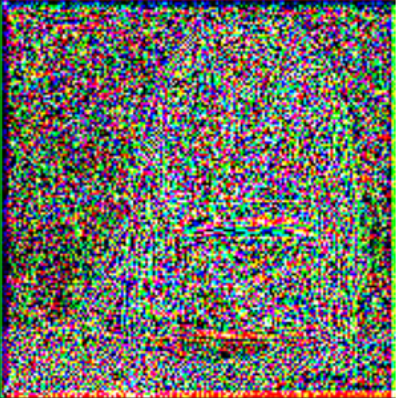}
        \caption{HH subband}
    \end{subfigure}
    \caption{Wavelet transformed components of a natural image. LL and HH subband capture the low-frequency and high frequency details.}
    \vspace{-0.4cm}
    \label{fig:bag_wavelet}
\end{figure}

% Based on the property, we propose the Kurtosis Concentration loss (KC loss).
Suppose, we need to train or finetune a diffusion model ${f_{\theta}}$ from input training images ($\{x\}$) with or without a conditioning vector $c$. The conditioning vector could be text, image, or none (in case of the unconditional diffusion model). 
The generated images obtained from $f_{\theta}$ given an initial noise map $\epsilon \sim N(0,I)$, a conditioning vector $c$ is given by $x_{gen} = f_{\theta} (x, \epsilon, c)$.
\textcolor{blue}{Typically, the diffusion model is trained to minimize the $l2$ distance between the ground truth image ($x$) and the noisy image ($x_{gen}$)~\cite{dhariwal2021diffusion} or their corresponding latent in case of Latent Diffusion Model (LDM)~\cite{rombach2022high}. Without loss of generality, we are referring that as reconstruction loss ($L_{recon}$) between the ground-truth image ($x$) and the generated image ($x_{gen}$), denoted by,
% Typically, the diffusion model is trained to minimize the reconstruction loss ($L_{recon}$) between the ground-truth image ($x$) and the generated image ($x_{gen}$), denoted by,
\begin{equation}
L_{recon} = \mathbb{E}_{x,c, \epsilon} [ \ || x_{gen} - x ||_{2}^{2}]  
\end{equation}}
\textcolor{blue}{Note that for LDM, this will be the $l2$ distance between the corresponding latents.}
Now, we will describe the KC loss. Note, that the KC property holds across different bandpass transformed domains (DCT, DWT, fastICA) and we choose DWT because it is widely used due to its hierarchical structure and energy compaction properties~\cite{e2008digital}.  
Typically, DWT transforms images into LL (low-low), LH (low-high), HL (high-low), HH (high-high) frequency bands and each of the sub-bands contains several sparse details of the image. E.g., LL and HH subband contains a low-pass and high-pass filtered version of the image respectively~\cite{zhang2014using} as shown in Fig.~\ref{fig:bag_wavelet}.
The generated image $x_{gen}$ is then transformed using Discrete Wavelet Transform (DWT) with kernels $k_1, k_2, .., k_n$ producing filtered images $g_{gen, 1}, g_{gen, 2}, .., g_{gen, n}$ respectively, such that, $g_{gen, i} = F_{k_i} (x_{gen})$.
Here, $F_{l}$ denotes the discrete wavelet transform with kernel $l$.

% Note that, kurtosis concentration is valid for FastICA, DCT, DWT etc, but we have chosen DWT because of its hierarchical structure and energy compaction properties~\cite{e2008digital}.

Now, kurtosis values of these $g_{gen, i}$ should be constant by the kurtosis concentration property, therefore, we minimize the difference between the maximum and minimum values of the kurtosis of $g_{gen, i}$'s to finetune the model using the loss, 
\begin{equation}
L_{KC} = \mathbb{E}_{x, c, \epsilon} [ ( \text{max}(\kappa(\{ g_{gen, i}\})) - \text{min}(\kappa(\{ g_{gen, i}\}))  )] 
\end{equation}

Here, $\kappa(x)$ is kurtosis of $x$.
Note that, this loss is quite generic and can be applied to both image or latent diffusion models for training. In the case of latent diffusion models, we need to transform the latent to image space, before applying this loss.  
In case of applying this loss to any task $T$ (DreamBooth, super-resolution, unconditional image generation), the overall loss ($L$) function would be, $L = L_{task} + L_{recon} + L_{KC}$, where $L_{task}$ is the task-specific loss.

% Notice that, the reconstruction and prior preservation losses are applied to latent space, since they are using latent diffusion model. However, this property holds for image space. Therefore, we need to transform the latent to image space, before applying the loss. 

%The overall loss function would be, $L = L_{recon} + L_{prior} + L_{KC}$ 

% \begin{equation}
% L = L_{recon} + L_{prior} + L_{KC}
% \end{equation}

%\subsubsection{Training details}

% \textbf{Training details.} We finetune the model with a batch-size of 8, using the AdamW optimizer with learning rate $5e-6$, for 800 steps. Using the pretrained diffusion model, we generate 10 images per class a priori to the training. More training details will be provided in the supplementary material.  
 
\section{Experiments}

We evaluate the efficacy of the proposed loss for three tasks - (1) personalized few-shot finetuning of diffusion model using text guidance, (2) unconditional image generation, and (3) image super-resolution.

\subsection{Task 1: Personalized few-shot finetuning using text guidance}

% \begin{figure}
%     \centering
%     \includegraphics[scale=0.4]{figures/problem.png}
%     \caption{\small{Personalized finetuning of text-to-image diffusion model. Given few training images of a dog backpack ([V] denotes the unique identifier), our task is to generate variations of that particular subject ([V] dog backpack) in diverse conditions given by the text-prompt (e.g., `` A [V] dog backpack on the beach'' as shown in figure).}} 
%     \label{fig:problem}
% \end{figure}

In this section, we address the problem of finetuning the text-to-image diffusion model from a few examples for text-guided image generation in a subject-driven manner. Specifically, given only a few images (e.g., 3-5) of a particular subject without any textual description, our task is to learn the subject-specific details and generate new images of that particular subject in different conditions specified by the text prompt. Suppose we are given four samples of a dog backpack. Now the task is to finetune the text-to-image diffusion model given these four samples of the particular dog backpack such that it learns the concept/subject etc. During inference, the model has to generate images containing that particular dog backpack according to the text prompt.

To evaluate the efficacy of KC loss in this task, we build upon two popular methods, (1) Dreambooth~\cite{ruiz2022dreambooth}, and (2) Custom diffusion~\cite{kumari2022multi}. In particular, we add KC loss to these frameworks while finetuning the denoising UNet to check whether the image quality improves and demonstrate the quality of generated images improves. 

% \begin{wrapfigure}{r}{0.8\textwidth}
% %\begin{figure}
%     \centering
%     \includegraphics[scale=0.25]{figures/db_cd_compare.jpg}
%     \caption{\small{Comparison of Dreambooth, Custom diffusion with/without KC loss. Adding KC loss improves image quality for both Dreambooth and Custom diffusion, in terms of color vividness, contrast, and lighting consistency.}}
%     \label{fig:db_cd_compare}
% %\end{figure}
% \end{wrapfigure}

%\begin{wrapfigure}{r}{0.8\textwidth}
\begin{figure}
    \centering
    \includegraphics[scale=0.24]{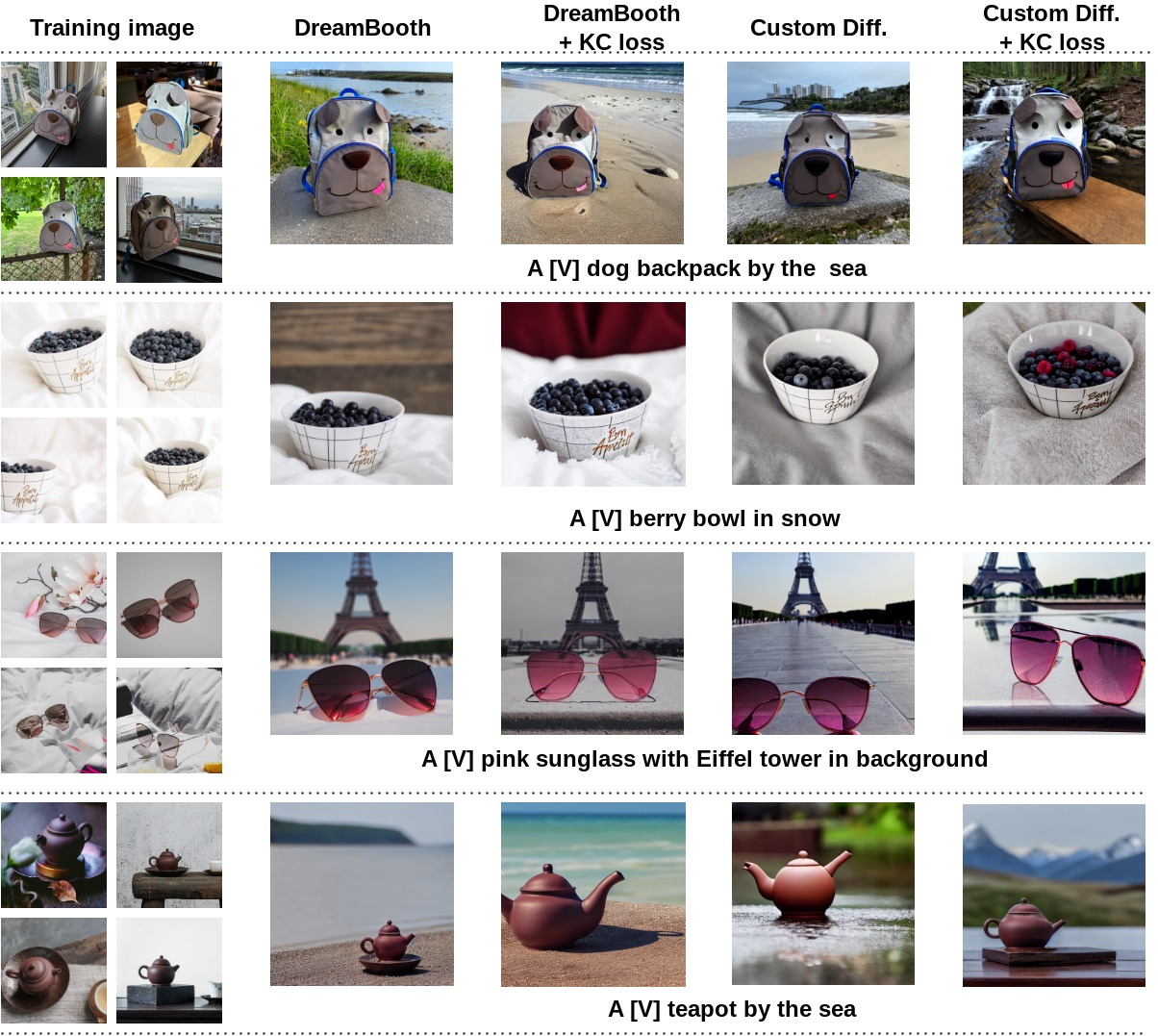}
    \caption{\small{Comparison of DreamBooth, Custom diffusion with/without KC loss. Adding KC loss improves image quality for both DreamBooth and Custom diffusion, in terms of color vividness, contrast, and lighting consistency.}}
    \vspace{-20pt}
    \label{fig:db_cd_compare}
\end{figure}
%\end{wrapfigure}

\textbf{Dataset and Metric.} We follow the dataset and experimental setup used by DreamBooth~\cite{ruiz2022dreambooth}. 
To evaluate the generated image quality with respect to the input image and the text prompt, we also use the subject fidelity metrics proposed by DreamBooth - (1) DINO, (2) CLIP-I, and (3) CLIP-T.
\textcolor{blue}{We have also compared with another naturalness loss, i.e., LPIPS loss~\cite{zhang2018unreasonable} as a baseline.}
% The dataset contains 30 subjects (e.g., backpack, stuffed animal, dogs, cats, sunglasses, cartoons etc) and 25 prompts including 20 re-contextualization prompts and 5 property modification prompts. 
% DINO, which is the average pairwise cosine similarity between the ViT-S/16 DINO embeddings~\cite{caron2021emerging} of the generated and real images. (2) CLIP-I, i.e., the average pairwise cosine similarity between CLIP~\cite{radford2015unsupervised} embeddings of the generated and real images. 
% To measure the prompt fidelity, we use CLIP-T, which is the average cosine similarity between prompt and image CLIP embeddings.

DreamBooth~\cite{ruiz2022dreambooth} finetunes the stable diffusion model using the standard reconstruction loss and a prior preservation loss.  However, it's prone to overfitting and some unnatural artifacts can be observed as shown in Fig.~\ref{fig:diffnat_teaser}.
For faster and lightweight training, custom diffusion~\cite{kumari2022multi} finetunes only the cross-attention module of the text-to-image stable diffusion model. 
% However, unnatural artefacts can be observed in the generated images as well as shown in Fig.~\ref{fig:db_cd_compare}. 
We evaluate both approaches with/without KC loss on the same dataset for a fair comparison.
When adding the proposed KC loss to these approaches, we obtain performance improvements in visual quality, i.e., FID~\cite{lucic2018gans}, MUSIQ score~\cite{ke2021musiq} as shown in Tab.~\ref{table:db_cd_diffnat_compare}. The qualitative results are shown in Fig.~\ref{fig:db_cd_compare}.
We follow the same setup for the dreambooth and custom diffusion baselines. Additionally, for KC loss, we decompose the reconstructed images using 27 `Daubechies' filter banks, and get the average deviation of the kurtosis values as a loss function. 
More training details will be provided in the supplementary material.

% We finetune the model with a batch-size of 8, using the AdamW optimizer with learning rate $5e-6$, for 800 steps. Using the pretrained diffusion model, we generate 10 images per class a priori to the training. More training details will be provided in the supplementary material.  

% %%%%% table in submission %%%%%%%%%%%%
% \begin{table}[!h]
% \small
% \centering
% \caption{Comparison of Personalized few-shot finetuning task}
% \scalebox{0.8}{
% \begin{tabular}{lcccccc}
% \toprule
% \multirow{2}{*}{Method}   		  & \multicolumn{2}{c}{\textbf{Image quality}}	& \multicolumn{2}{c}{\textbf{Subject fidelity}} & \multicolumn{1}{c}{\textbf{Prompt fidelity}}\\
% \cmidrule(lr){2-3} \cmidrule(lr){4-5} \cmidrule(lr){6-6}
% & {\bf FID score $\downarrow$} & {\bf MUSIQ score $\uparrow$} & {\bf DINO $\uparrow$}  & {\bf CLIP-I $\uparrow$} & {\bf CLIP-T $\uparrow$} \\
% \midrule 
% {DreamBooth~\cite{ruiz2022dreambooth}} & {111.76} & {68.319} & {0.65} & {0.81} & {0.31} \\
% {DreamBooth + KC loss(Ours)} & {\bf 100.08} & {\bf 69.78} & {\bf 0.68 } & {\bf 0.84} & {\bf 0.34} \\
% \hline
% {Custom Diff.~\cite{kumari2022multi}} & {84.65} & {70.15} & {0.71} & {0.87} & {0.38} \\
% {Custom Diff. + KC loss(Ours)} & {\bf 75.68} & {\bf 72.22} & {\bf 0.73 } & {\bf 0.88} & {\bf 0.40} \\
% \bottomrule
% \end{tabular}}
% \label{table:db_cd_diffnat_compare}
% \end{table}
% %%%%%%%%%%%%%%%%%%%%%%%%%%%%%%%%%%%%%%

%%%% added in rebuttal %%%%%%%%%%%%%%%
\begin{table}[!h]
\small
\centering
\vspace{-0.2cm}
\caption{\textcolor{blue}{Comparison of Personalized few-shot finetuning task}}
\vspace{-0.2cm}
\scalebox{0.85}{
\begin{tabular}{lcccccc}
\toprule
\multirow{2}{*}{Method}   		  & \multicolumn{2}{c}{\textbf{Image quality}}	& \multicolumn{2}{c}{\textbf{Subject fidelity}} & \multicolumn{1}{c}{\textbf{Prompt fidelity}}\\
\cmidrule(lr){2-3} \cmidrule(lr){4-5} \cmidrule(lr){6-6}
& {\bf FID score $\downarrow$} & {\bf MUSIQ score $\uparrow$} & {\bf DINO $\uparrow$}  & {\bf CLIP-I $\uparrow$} & {\bf CLIP-T $\uparrow$} \\
\midrule 
{DreamBooth~\cite{ruiz2022dreambooth}} & {111.76} & {68.31} & {0.65} & {0.81} & {0.31} \\
{DreamBooth~\cite{ruiz2022dreambooth} + LPIPS} & {108.23} & {68.39} & {0.65} & {0.80} & {0.32} \\
{DreamBooth + KC loss(Ours)} & {\bf 100.08} & {\bf 69.78} & {\bf 0.68 } & {\bf 0.84} & {\bf 0.34} \\
\hline
{Custom Diff.~\cite{kumari2022multi}} & {84.65} & {70.15} & {0.71} & {0.87} & {0.38} \\
{Custom Diff.~\cite{kumari2022multi} + LPIPS} & {80.12} & {70.56} & {0.71} & {0.87} & {0.37} \\
{Custom Diff. + KC loss(Ours)} & {\bf 75.68} & {\bf 72.22} & {\bf 0.73 } & {\bf 0.88} & {\bf 0.40} \\
\bottomrule
\end{tabular}}
\label{table:db_cd_diffnat_compare}
\end{table}
%%%%%%%%%%%%%%%%%%%%%%%%

\begin{wrapfigure}{r}{0.35\textwidth}
    \centering
    \includegraphics[width=0.3\textwidth]{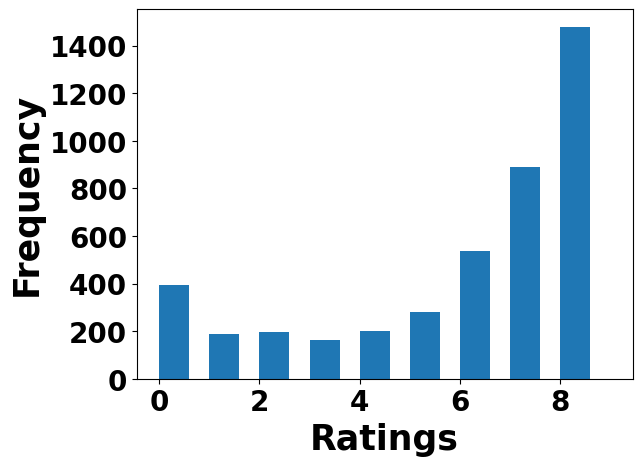}
    \vspace{-10pt}
    \caption{\small{Subject fidelity assessment by user study. The ratings ranges from ``0'' being ``extremely unlikely'' to 10 being ``extremely likely''. We observe from the plot that most of the users find DiffNat preserves subject fidelity. The average rating is 5.8, which is ``moderately likely'' to ``highly likely''.}} 
    \label{fig:user_study_1}
    \vspace{-20pt}
\end{wrapfigure}

\textbf{Human evaluation.} 
Since the perceptual quality is quite subjective, automatic metrics do not correlate well with the perceptual studies~\cite{zhang2018unreasonable}. To verify that the improved scores actually correspond to better quality images,
we evaluate our approach using human preference study through Amazon Mechanical Turk. Specifically, we performed two human evaluation tasks - (1) subject fidelity assessment and (2) image quality ranking. 

In the subject fidelity assessment, we conduct Two Alternative Forced choice (2AFC) experiment setup. In particular, we show a pair of images containing the real image and the edited image using KC loss and asked the user the question : ``How similar are these two objects?'', with 10 options ranging from ``extremely likely'' to ``extremely unlikely''( with ``0'' being ``extremely unlikely'' to 10 being ``extremely likely''). We test this with 423 samples with 10 human evaluations per sample, totaling 4230 tasks. We show the aggregate response in Fig.~\ref{fig:user_study_1}, which reveals that adding our proposed loss retains subject fidelity in most cases.

Next, we provide 30 examples of natural images and corresponding generated images using Dreambooth, Custom diffusion, and our method, and asked the question : ``which of the edited images is of best visual quality considering factors including image quality and preserving the identity of the original image?'' \textcolor{blue}{We evaluate this by 50 users, totaling 1500 questionnaires and the aggregate response reveals that DiffNat-generated images outperform the baselines by a large margin (i.e., 50.4\%,) where the available options are \{ `DiffNat', `Dreambooth', `Custom diffusion', `None is satisfactory'\}, which shows that our approach outperforms the baseline approaches.} 

\subsection{Task 2: Unconditional image generation}

Unconditional image generation does not require any text or image guidance. It simply tries to learn the training data distribution through a generative model (here we focus on the diffusion model) and generates samples similar to the training data distribution. Denoised Diffusion Probabilistic Model (DDPM)~\cite{ho2020denoising} is a parameterized Markov chain that is trained to generate matched data distribution through variational inference. The denoising takes place in the image space and produces better image quality compared to GANs. 

\begin{wrapfigure}{r}{0.5\textwidth}
%\begin{figure}
\vspace{-20pt}
    \centering
    \includegraphics[scale=0.3]{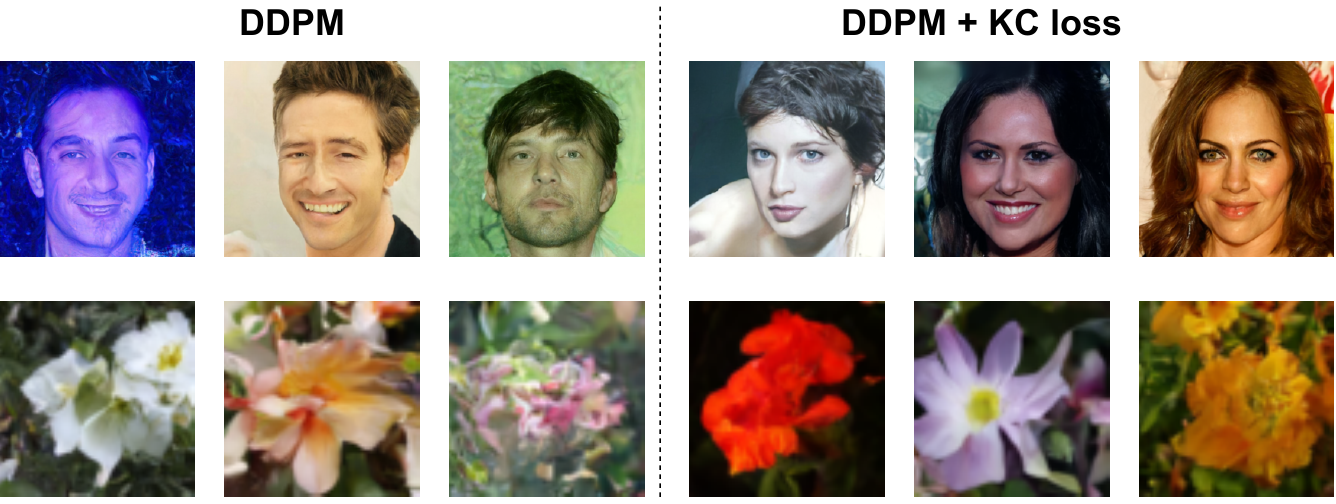}
    %\vspace{-10pt}
    \caption{Comparison of unconditional image generation (DDPM) with/without KC loss. Integrating KC loss significantly improve image quality, whereas DDPM generated images have unnatural image artifacts.} 
    \label{fig:uncond_compare}
    \vspace{-20pt}
%\end{figure}
\end{wrapfigure}

We incorporate our proposed KC loss in this framework and obtain even better perceptual quality on various diverse datasets in terms of FID and MUSIQ score. The experimental results are shown in Tab.~\ref{table:unconditional_image_gen} and Fig.~\ref{fig:uncond_compare}. Note that, in this approach, we integrate the KC loss directly into image space, which shows the flexibility of our proposed loss. We have experimented with Oxford-flowers~\cite{nilsback2006visual}, celebAfaces~\cite{zhang2020celeba} and CelebAHQ~\cite{karras2017progressive} datasets and obtained consistent improvements on image quality as shown in Tab.~\ref{table:unconditional_image_gen}. Qualitative analysis in Fig.~\ref{fig:uncond_compare} verify that integrating KC loss improves image quality in terms of details, contrast, and color vividness. 

% %%%%%% table in submission %%%%%%%%%%
% \begin{table}[!h]
% \small
% \centering
% \caption{Comparison of unconditional image generation task}
% \scalebox{0.7}{
% \begin{tabular}{lccccccc}
% \toprule
% \multirow{2}{*}{Method}   		  & \multicolumn{2}{c}{\textbf{Oxford flowers}}	& \multicolumn{2}{c}{\textbf{Celeb-faces}} & \multicolumn{2}{c}{\textbf{CelebAHQ}}\\
% \cmidrule(lr){2-3} \cmidrule(lr){4-5} \cmidrule(lr){6-7}
% & {\bf FID score $\downarrow$} & {\bf MUSIQ score $\uparrow$} & {\bf FID score $\downarrow$} & {\bf MUSIQ score $\uparrow$} & {\bf FID score $\downarrow$} & {\bf MUSIQ score $\uparrow$} \\
% \midrule 
% {DDPM~\cite{ruiz2022dreambooth}} & {243.43} & {20.67} & {202.67} & {19.07} & {199.77} & {46.05} \\
% {DDPM + KC loss(Ours)} & {\bf 237.73} & {\bf 21.13} & {\bf 198.23 } & {\bf 19.52} & {\bf 190.59} & {\bf 46.83} \\
% % {Custom Diff.~\cite{kumari2022multi}} & {84.65} & {70.15} & {-} & {-} & {-} \\
% % {Custom Diff. + KC loss(Ours)} & {\bf 75.68} & {\bf 72.22} & {\bf - } & {\bf -} & {\bf -} \\
% \bottomrule
% \end{tabular}}
% \label{table:unconditional_image_gen}
% \end{table}
% %%%%%%%%%%%%%%%%%%%%%%%%%%%%%%%%%%%%%%

%%%%%%%%%%% table in rebuttal %%%%%%%%%%
\begin{table*}[!h]
\small
\centering
\vspace{-0.2cm}
\caption{\textcolor{blue}{Comparison of unconditional image generation task}}
\vspace{-0.2cm}
\scalebox{0.8}{
\begin{tabular}{lccccccc}
\toprule
\multirow{2}{*}{Method}   		  & \multicolumn{2}{c}{\textbf{Oxford flowers}}	& \multicolumn{2}{c}{\textbf{Celeb-faces}} & \multicolumn{2}{c}{\textbf{CelebAHQ}}\\
\cmidrule(lr){2-3} \cmidrule(lr){4-5} \cmidrule(lr){6-7}
& {\bf FID score $\downarrow$} & {\bf MUSIQ score $\uparrow$} & {\bf FID score $\downarrow$} & {\bf MUSIQ score $\uparrow$} & {\bf FID score $\downarrow$} & {\bf MUSIQ score $\uparrow$} \\
\midrule 
{DDPM~\cite{ho2020denoising}} & {243.43} & {20.67} & {202.67} & {19.07} & {199.77} & {46.05} \\
{DDPM~\cite{ho2020denoising} + LPIPS} & {242.62} & {20.80} & {201.55} & {19.21} & {197.17} & {46.15} \\
{DDPM + KC loss(Ours)} & {\bf 237.73} & {\bf 21.13} & {\bf 198.23 } & {\bf 19.52} & {\bf 190.59} & {\bf 46.83} \\
% {Custom Diff.~\cite{kumari2022multi}} & {84.65} & {70.15} & {-} & {-} & {-} \\
% {Custom Diff. + KC loss(Ours)} & {\bf 75.68} & {\bf 72.22} & {\bf - } & {\bf -} & {\bf -} \\
\bottomrule
\end{tabular}}
\label{table:unconditional_image_gen}
\end{table*}
%%%%%%%%%%%%%%%%%%%%%%%%%%%%%%%%%%%%%%%%%%%%%%%

\begin{wrapfigure}{r}{0.5\textwidth}
    \vspace{-50pt}
    \centering
    \includegraphics[width=0.5\textwidth]{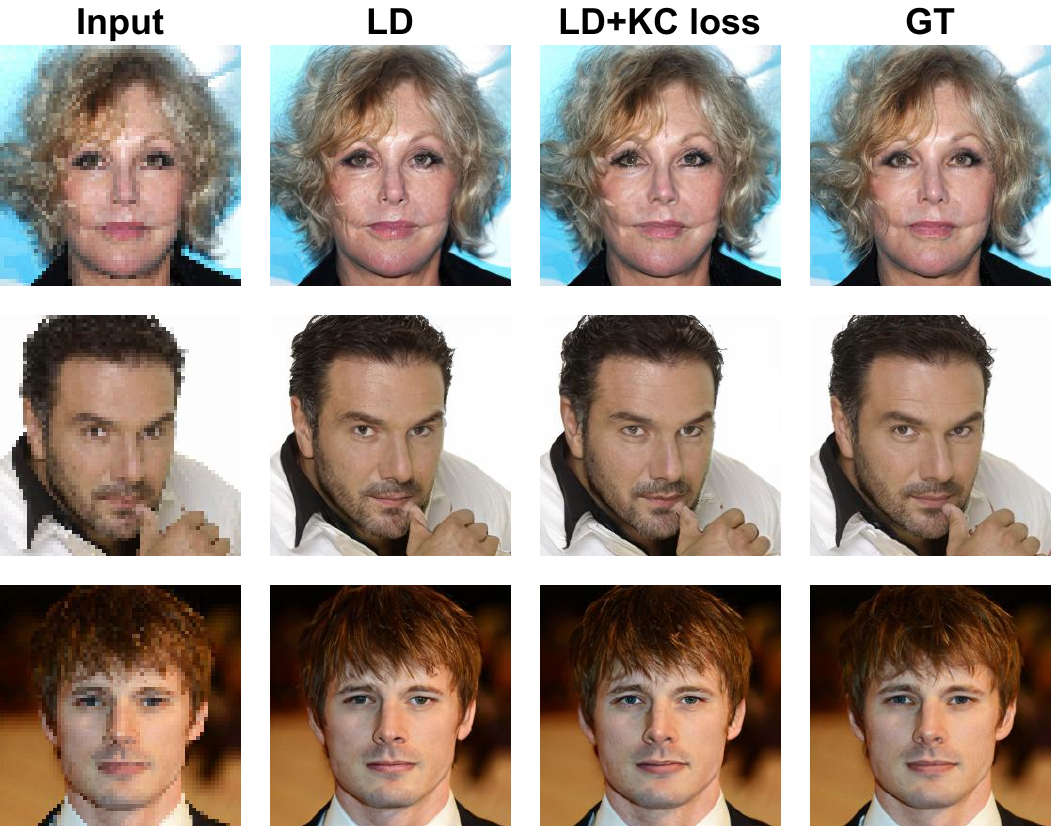}
    
    \caption{\small{Image super-resolution quality improves adding KC loss to k-diffusion (LD) framework}. Generated images show better quality in terms of overall image smoothness (first row), finer details like eyes (second and third row).} 
    \label{fig:super_resol_ld}
    \vspace{-20pt}
\end{wrapfigure}

\subsection{Task 3: Image super-resolution}
%%%%%%%%%%%%%%%%%%%%%%%%%%%%%
%%%%%%%%%%%%%%%%%%%%%%%%%%%%%
Image super-resolution typically takes the form of a conditional generation task, leveraging a low-resolution image as an additional condition for the diffusion model. In this study, we establish two state-of-the-art diffusion pipelines as baselines for comparison. Guided diffusion (GD)~\cite{dhariwal2021diffusion} directly takes the low-resolution image as a condition and performs the diffusion operation in the pixel space. Additionally, we also explore the latent diffusion model (LDM)~\cite{rombach2022high} that operates in the latent space of a pre-trained VQVAE~\cite{esser2021taming}. We introduce conditioning by utilizing the latent embedding of the low-resolution image with this model, referring to it as conditional-LDM (cLDM).

Note that, as GD operates in the pixel space, we directly add the proposed KC loss to the output of the denoising UNet. Conversely, for cLDM, we initially convert the latent embedding to image space using the pre-trained decoder and integrate the KC loss on the output of the decoder. For training, we use the standard FFHQ dataset~\cite{karras2017progressive}, which contains 70k high-quality images. Specifically, we address the task of $\times 4$ super-resolution where the GT images are of resolution $256 \times 256$. We evaluate randomly sampled 1000 images from CelebA-Test dataset \cite{karras2017progressive} under the same $\times 4$-SR setting.

% %%%%%%%% table in submission %%%%%%%%%%%%%%%%%%%%
% \begin{wraptable}{r}{0.7\textwidth}
% %\begin{table}[!h]
% \small
% \centering
% \vspace{-10pt}
% \caption{Comparison of image super-resolution task}
% \scalebox{0.7}{
% \begin{tabular}{lcccccc}
% \toprule
% \multirow{2}{*}{Method}   		  & \multicolumn{5}{c}{\textbf{Image quality}}\\
% \cmidrule(lr){2-6} 
% & {\bf FID score $\downarrow$} & {\bf PSNR $\uparrow$} & {\bf SSIM $\uparrow$}  & {\bf LPIPS $\downarrow$} & {\bf MUSIQ score $\uparrow$} \\
% \midrule 
% {GD~\cite{dhariwal2021diffusion}} & {121.23} & {18.13} & {0.54} & {0.28} & {57.31} \\
% {GD + KC loss(Ours)} & {\bf 103.19} & {\bf 18.92} & {\bf 0.55 } & {\bf 0.26} & {\bf 58.69} \\
% \hline
% {LD.~\cite{karras2022elucidating}} & {95.83} & {19.16} & {0.56} & {0.257} & {59.57} \\
% {LD + KC loss(Ours)} & {\bf 83.34} & {\bf 20.25} & {\bf 0.58 } & {\bf 0.221} & {\bf 61.20} \\
% \bottomrule
% \end{tabular}
% \label{table:compare_super_resolution}}
% \vspace{-10pt}
% %\end{table}
% \end{wraptable}
% %%%%%%%%%%%%%%%%%%%%%%%%%%%%%%%%%%%%%%%%%%%%%%%%%

%%%% table in rebuttal %%%%%%%%%%%%%%%%%%%%%%
%\begin{table}
%%\begin{table}[!h]
\begin{wraptable}{r}{0.7\textwidth}
\small
\centering
\vspace{-0.5cm}
\caption{\textcolor{blue}{Comparison of image super-resolution task}}
\vspace{-0.2cm}
\scalebox{0.7}{
\begin{tabular}{lcccccc}
\toprule
\multirow{2}{*}{Method}   		  & \multicolumn{5}{c}{\textbf{Image quality}}\\
\cmidrule(lr){2-6} 
& {\bf FID score $\downarrow$} & {\bf PSNR $\uparrow$} & {\bf SSIM $\uparrow$}  & {\bf LPIPS $\downarrow$} & {\bf MUSIQ score $\uparrow$} \\
\midrule 
{GD~\cite{dhariwal2021diffusion}} & {121.23} & {18.13} & {0.54} & {0.28} & {57.31} \\
{GD~\cite{dhariwal2021diffusion} + LPIPS} & {119.81} & {18.22} & {0.54} & {0.27} & {57.42} \\
{GD + KC loss(Ours)} & {\bf 103.19} & {\bf 18.92} & {\bf 0.55 } & {\bf 0.26} & {\bf 58.69} \\
\hline
{LD.~\cite{karras2022elucidating}} & {95.83} & {19.16} & {0.56} & {0.26} & {59.57} \\
{LD.~\cite{karras2022elucidating} + LPIPS} & {92.77} & {19.42} & {0.57} & {0.25} & {59.82} \\
{LD + KC loss(Ours)} & {\bf 83.34} & {\bf 20.25} & {\bf 0.58 } & {\bf 0.22} & {\bf 61.20} \\
\bottomrule
\end{tabular}
\label{table:compare_super_resolution}}
\vspace{-10pt}
%%\end{table}
%\end{table}
\end{wraptable}
%%%%%%%%%%%%%%%%%%%%%%%%%%%%%%%%%%%%%%%%%%%%%

\begin{wrapfigure}{r}{0.5\textwidth}
    \centering
    \vspace{-10pt}
    \includegraphics[width=0.5\textwidth]{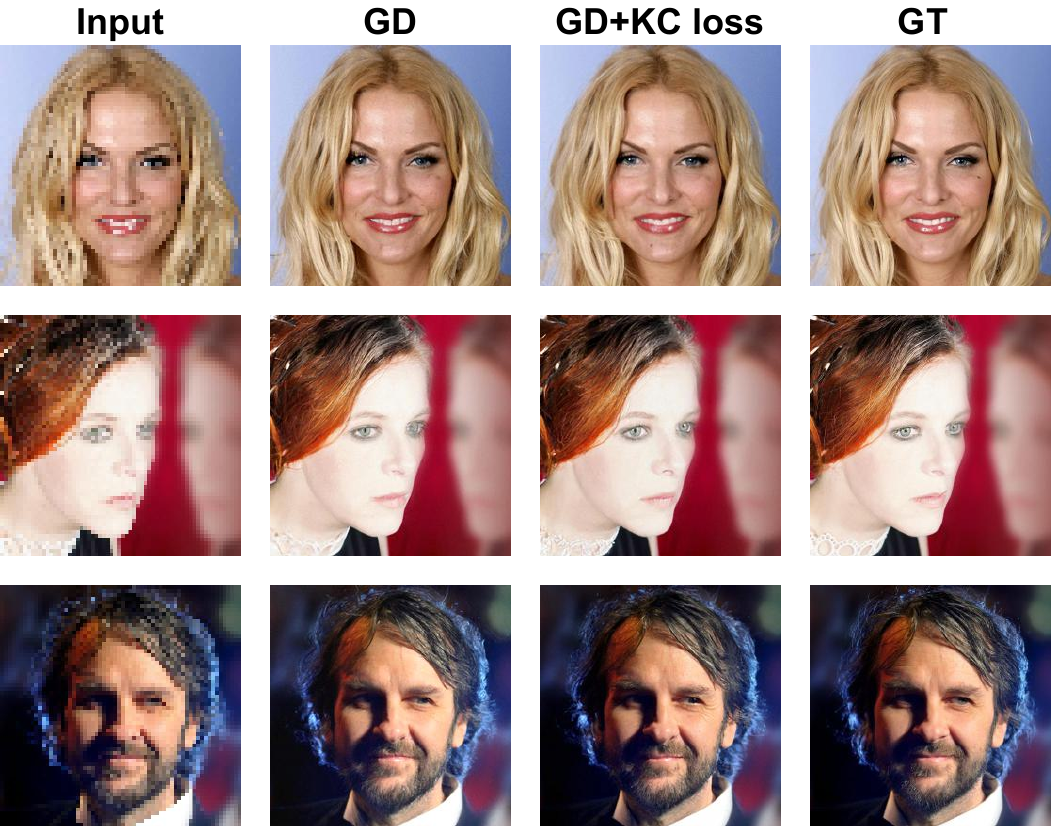}
    % \vspace{-10pt}
    \caption{\small{Image super-resolution quality improves adding KC loss to the guided diffusion (GD) framework}. Generated images show better quality in terms of overall image details (first row), finer details like eyes (second row), and overall color and brightness (third row).} 
    \label{fig:super_resol_gd}
    \vspace{-20pt}
\end{wrapfigure}

Since, the proposed KC loss improves image quality, it is inherently applicable for this task. We integrate KC loss in the SOTA super-resolution diffusion models~\cite{dhariwal2021diffusion, rombach2022high}, and obtained performance improvement in perceptual quality as shown in Tab.~\ref{table:compare_super_resolution}. 
In the qualitative results shown in Fig.~\ref{fig:super_resol_gd} and Fig.~\ref{fig:super_resol_ld}, we observe that adding KC loss improves the image quality and finer details, e.g., eye structure, texture, lighting etc.

\subsection{Comparison of real vs synthetic detection}

To perform the robustness analysis of the proposed loss, we also perform the following experiment. We train a classifier (2-layer MLP on top of pre-trained ResNet feature extractor) to distinguish real vs synthetic, where `real' comes from natural image belongs to the DreamBooth dataset and `synthetic' comes from diffusion model generated images from algorithm X or X + KC loss. Here `X' can be `DreamBooth' or `Custom diffusin''. For testing, we select non-overlapping test samples for both natural and diffusion generated images. 
When tested on DreamBooth and Custom diffusion, we observe that adding KC loss decrease the real vs synthetic classification accuracy as shown in Tab.~\ref{table:real_fake}. This indicates that generated images are of superior perceptual quality, exhibiting a greater degree of ``naturalness'' to both human observers and machine algorithms alike.
 
% Hence, adding KC loss during the training of the diffusion model leads to generated images of superior perceptual quality, exhibiting a greater degree of ``naturalness'' to both human observers and machine algorithms alike.
% Hence, the diffusion generated images, adding KC loss looks more natural to the human observer and the machine both.

\begin{wraptable}{r}{0.5\textwidth}
%\begin{table}[!h]
  %\scriptsize
  \small
  \centering
  \vspace{-10pt}
  \caption{Comparison of real vs synthetic detection}
  \scalebox{0.8}{
  %\caption{Comparison of real vs fake detection}
  \begin{tabular}{lcc}
  \toprule
  {\bf Method} & {\bf Accuracy} \\ 
  \midrule
  {DreamBooth~\cite{ruiz2022dreambooth}} & {93.33\%} \\
  {DreamBooth~\cite{ruiz2022dreambooth} + KC loss} & {66.66\%} \\  
  \hline
  {Custom Diffusion~\cite{kumari2022multi}} & {94.16 \%}\\  
  {Custom Diffusion~\cite{kumari2022multi} + KC loss} & {92.5\%} \\
  \bottomrule
  \end{tabular}}
  \label{table:real_fake}
  \vspace{-10pt}
 %\end{table}
\end{wraptable}

\section{Conclusion}

Although diffusion models have significantly advanced in creating naturalistic images, these images can have unnatural artifacts, especially in the cases of few-shot finetuning of large-scale text-to-image diffusion models.
We leverage the kurtosis concentration property of natural images to define a novel and generic loss function in order to preserve the ``naturalness'' of generated images. Kurtosis concentration property suggests that the kurtosis values across different bandpass versions of the natural image tend to be constant. The proposed kurtosis concentration loss minimizes the gap between the maximum and minimum value of the kurtosis across different DWT filtered versions of the image.
We show this loss improves image quality for diverse generative tasks - (1) personalized few-shot finetuning of text-to-image diffusion model, (2) unconditional image generation, and (3) image super-resolution. We also conduct human studies to validate our approach.

% We address the problem of ``personalization'' of text-to-image diffusion model, where given few training examples of a particular identity we have to finetune the text-to-image diffusion model to generate variant of that particular identity in diverse scenes, views, lighting condition etc. To this end, we propose DiffNat, a finetuning method designed to improve the perceptual quality of generated image. Our motivation comes from the kurtosis concentration property of natural images, which says the kurtosis values across different band-pass (DWT) version of images tend to be a constant value. Experimental results both qualitatively and quantitatively validate our approach.  

% \section{Acknowledgement}
% \label{sec:conclusion}
% %\vspace{-0.2cm}
% The authors AR, AS and RC are supported by an ONR MURI grant N00014- 20-1-2787. 

%\newpage

\bibliography{iclr2024_conference}
\bibliographystyle{iclr2024_conference}

\appendix
\section{Appendix}

%\section{Introduction}

In this supplementary material, we will provide the following details. 
\begin{enumerate}
    \item Training details.
    \item Theoretical justification.
    \item Additional experimental results.
    \item Failure cases.
    \item \textcolor{blue}{Computation complexity}
    \item \textcolor{blue}{Training time analysis}
    \item \textcolor{blue}{Kurtosis analysis}
    \item \textcolor{blue}{Convergence analysis}
    \item \textcolor{blue}{Qualitative analyis}
\end{enumerate}

\section{Training details}

The training details of finetuning the diffusion model for various tasks have been provided here. 
For personalized few-shot finetuning, we consider two methods - Dreambooth~\cite{ruiz2022dreambooth} and Custom diffusion~\cite{kumari2022multi}. For fair comparison, we applied both the approaches on the dataset and setting introduced by Dreambooth. 
The dataset contains 30 subjects (e.g., backpack, stuffed animal, dogs, cats, sunglasses, cartoons etc) and 25 prompts including 20 re-contextualization prompts and 5 property modification prompts. 
DINO, which is the average pairwise cosine similarity between the ViT-S/16 DINO embeddings~\cite{caron2021emerging} of the generated and real images. (2) CLIP-I, i.e., the average pairwise cosine similarity between CLIP~\cite{radford2015unsupervised} embeddings of the generated and real images. To measure the prompt fidelity, we use CLIP-T, which is the average cosine similarity between prompt and image CLIP embeddings.

For unconditional image generation, we have experimented on oxford flowers, CelebAfaces and CelebAHQ datasets. Image quality has been measured by FID and MUSIQ score.

In case of image super-resolution, we experimented with guided diffusion~\cite{dhariwal2021diffusion} and latent diffusion~\cite{karras2022elucidating} pipelines. We use FFHQ dataset for training, and test on a subset of 1000 images from CelebAHQ test set for x4 super-resolution task. 
The hyperparameter details are given in Tab.~\ref{tab:diffnat_hparams}.

\begin{table}[!h]
    \centering
    \caption{Hyperparameters}
    \scalebox{0.85}{
    \begin{tabular}{cc}
    \toprule
    Hyperparameter & Values \\
    \midrule
    Coefficient of $L_{recon}$ & 1 \\
    Coefficient of $L_{prior}$ & 1 \\
    Coefficient of $L_{KC}$ & 1 \\
    Learning rate & $10^{-5}$\\
    Batch size (Dreambooth, Custom diffusion) & 8\\
    Batch size (DDPM) & 125\\
    Batch size (GD) & 16 \\
    Batch size (LD) & 9 \\
    Text-to-image diffusion model & Stable Diffusion-v1~\cite{rombach2022high} \\
    Number of class prior images (Dreambooth, Custom diffusion ) &  10 \\
    Number of DWT components & 25\\
    \bottomrule
    \end{tabular}}
    \label{tab:diffnat_hparams}
\end{table}

\section{Theoretical justification}

Here we provide theoretical analysis of the Lemmas mentioned in the main paper.
\begin{lemma}
\label{lemma: lemma2}
A Gaussian scale mixture (GSM) vector $x$ with zero mean has the following probability density function:
\begin{equation}
    p(x) = \int_{0}^{\infty} \mathcal{N}(x;0,z\Sigma_{x})p_{z}(z)dz
\end{equation}
and its projection kurtosis is \underline{constant} with respect to the projection direction w, i.e.,
\begin{equation}
    \kappa(w^Tx) = \frac{3var_{z}\{z\}}{\mathcal{E}_{z}\{z\}^2}
\end{equation}
where $\mathcal{E}_{z}\{z\}$ and $var_{z}\{z\}$ are the mean and variance of latent variable $z$ respectively.
\end{lemma}

\textit{Proof.} 
Marginal distribution of the projection of $x$ on non-zero vector $w$ is given by,
\begin{align*}
    p_{w}(t) & = \int_{x: w^Tx=t} p(x)dx \\
             & = \int_{z} p_{z}(z)dz. \int_{x:w^Tx=t} \frac{1}{\sqrt{(2\pi z)^d} |det(\Sigma_{x})|} exp(-\frac{x^T\Sigma_{x}^{-1}x}{2z})dx \\
             & = \int_{z} \mathcal{N}_{t}(0, zw^T\Sigma_{x}w)p_{z}(z)dz \\
\end{align*}

Note that, the last equality holds from the marginalization property of Gaussian, i.e., $X \approx \mathcal{N}(\mu, \Sigma)$, then, $AX \approx \mathcal{N}(A\mu, A \Sigma A^T)$.

The variance of $w^Tx$,
\begin{align*}
    \mathcal{E}_{t}\{t^2\} & = \int_{z} p_{z}dz \int_{t} t^2 \mathcal{N}_{t}(0, zw^T\Sigma_{x}w)dz \\
                            & = w^T\Sigma_{x}w \int_{z} z p_{z}dz \\
                            & = w^T\Sigma_{x}w \mathcal{E}_{z} \{z\}
\end{align*}

The fourth order moment of $w^Tx$,
\begin{align*}
    \mathcal{E}_{t}\{t^4\} & = \int_{z} p_{z}dz \int_{t} t^4 \mathcal{N}_{t}(0, zw^T\Sigma_{x}w)dz \\
                            & = 3 (w^T\Sigma_{x}w)^2 \int_{z} z^2 p_{z}dz \\
                            & = 3 (w^T\Sigma_{x}w)^2 \mathcal{E}_{z} \{z^2\}
\end{align*}
We utilize the property that $\mathcal{N}_{t}(0,\sigma^2)$ has a fourth order moment of $3 \sigma^4$.

Finally, the kurtosis becomes,
\begin{align*}
    \kappa(w^Tx) & =  \frac{ \mathcal{E}_{t}\{t\}^4 }{ \mathcal{E}_{t}\{t\}^2 } - 3 \\
                 & =  \frac{ 3 \mathcal{E}_{z}\{z\}^2 }{ \mathcal{E}_{z}\{z\}^2 } - 3 \\
                 & =  \frac{ 3 ( \mathcal{E}_{z}\{z^2\} - \mathcal{E}_{z}\{z\}^2 ) }{ \mathcal{E}_{z}\{z\}^2 }\\
                  & = \frac{3var_{z}\{z\}}{\mathcal{E}_{z}\{z\}^2}
\end{align*}

\begin{lemma}
\label{lemma: lemma3}
  If the noisy version of the natural image is denoted by, y = x + n, where x is a whitened GSM vector (normalized natural image) and n is a zero-mean white Gaussian noise with variance $\sigma^2I$, x and n are mutually independent of each other, then the projection kurtosis of y, $\kappa(w^Ty)$ can be expressed as:
\begin{equation}
    \kappa(w^Ty) = \kappa(w^Tx) \Big(1 - \frac{c}{SNR(y)} \Big)^2 = \frac{3var_{z}\{z\}}{\mathcal{E}_{z}\{z\}^2} \Big(1 - \frac{c}{SNR(y)} \Big)^2
\end{equation}

where Signal-to-Noise Ratio (SNR) is defined as, $SNR(y) = \frac{\sigma^2(y)}{\sigma^2(n)}$ and $c$ is a constant.
\end{lemma}

\textit{Proof.}
Here, we provide the proof of Lemma 1, mentioned in the main paper.
Without loss of generality, we strat by assuming, $\mathcal{E}_{x}{x} = 0$, since the mean can be easily subtracted from the data. 
We also assume that n is a zero-mean white Gaussian noise with variance $\sigma^2I$, x and n are mutually independent of each other. 

\begin{align*}
    \sigma^2(w^Tn) & = w^T \mathcal{E}_{z} \{zz^T\} w = \sigma^2w^Tw = \sigma^2 \\
    \sigma^2(w^Tx) & = w^T \mathcal{E}_{x} \{xx^T\} w = w^T \Sigma_{x} w\\
    \sigma^2(w^Ty) & = \sigma^2(w^xy) + \sigma^2(w^Tn) = w^T \Sigma_{x} w + \sigma^2 \\
\end{align*}

Since $n$ is a white Gaussian, $x$ and $n$ are independent, then $w^Tx$ and $w^Tn$ 
Therefore,
\begin{equation}
    \sigma^2(w^Ty) = \sigma^2(w^Tx) + \sigma^2(w^Tn)
\end{equation}.

Similarly, for fourth order moment, using the additivity of cumulants of independent variables (since $x$ and $n$ are independent)~\cite{papoulis2002probability}, we obtain, 

\begin{equation}
\begin{split}
    \kappa(w^Ty)(\sigma^2(w^Ty))^2 & = \kappa(w^Tx)(\sigma^2(w^Tx))^2 + \kappa(w^Tn) (\sigma^2(w^Tn))^2 \\
    & = \kappa(w^Tx)(\sigma^2(w^Tx))^2 \\
\end{split}
\end{equation}
Since, For Gaussian, $\kappa(n) = 0$

By rearranging, we have, 
\begin{align*}
    \kappa(w^Ty)& = \kappa(w^Tx). \big( \frac{\sigma^2(w^Tx)}{\sigma^2(w^Ty)} \big)^2 \\
                & = \kappa(w^Tx). \big( \frac{\sigma^2(w^Ty) - \sigma^2 }{\sigma^2(w^Ty)} \big)^2 \\
                & = \kappa(w^Tx). \big( 1 - c.\frac{\sigma^2}{\sigma^2(y)} \big)^2 \\
                & = \frac{3var_{z}\{z\}}{\mathcal{E}_{z}\{z\}^2}. \big( 1 - \frac{c}{SNR(y)} \big)^2 \\
\end{align*}
    
Here, Signal-to-Noise Ratio (SNR) is defined as, $\text{SNR}(y) = \frac{\sigma^2(y)}{\sigma^2(n)}$.

\begin{figure*}
    \centering
    \includegraphics[scale=0.4]{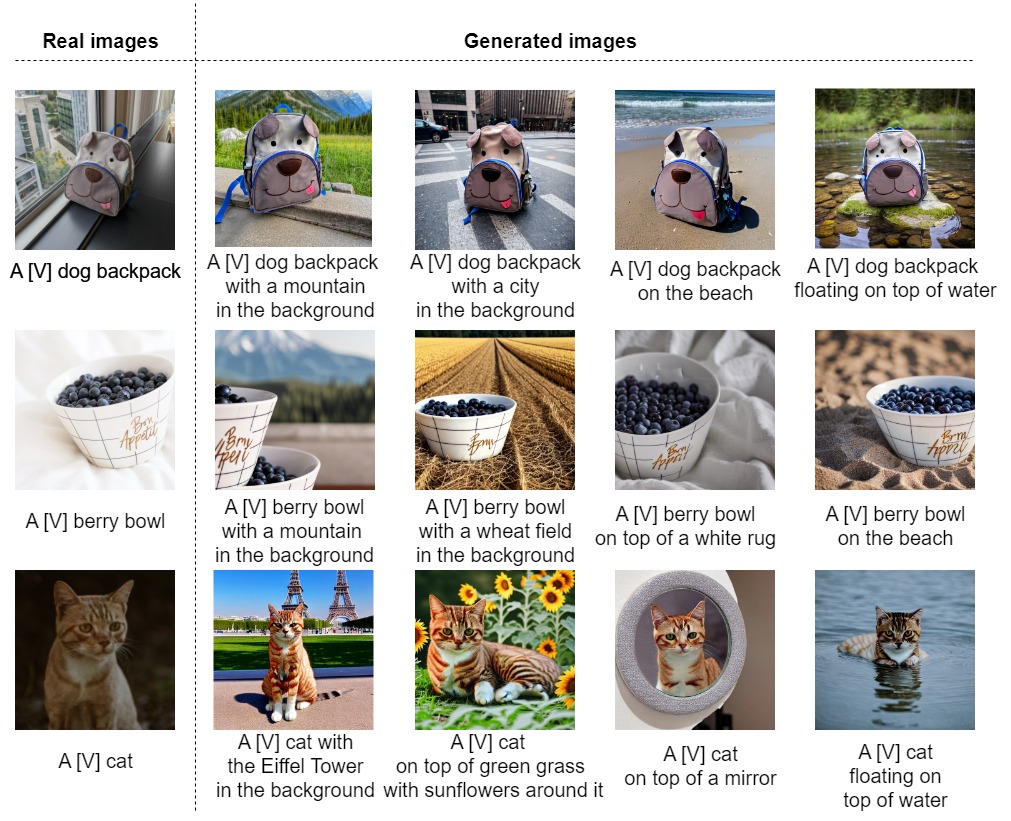}
    \caption{\textcolor{blue}{DiffNat generated images. The task is to learn a unique identifier (``A [V] dog backpack'') of the training images and generate variations w.r.t. background, lighting conditions etc. The generated images look natural in different background context, e.g., ``A [V] dog backpack on the beach/ with a city in the background etc''. The generated images are of high quality.}}
    \label{fig:diffnat_exp}
\end{figure*}

\section{Additional experimental results}

\begin{figure*}
    \centering
    \includegraphics[scale=0.65]{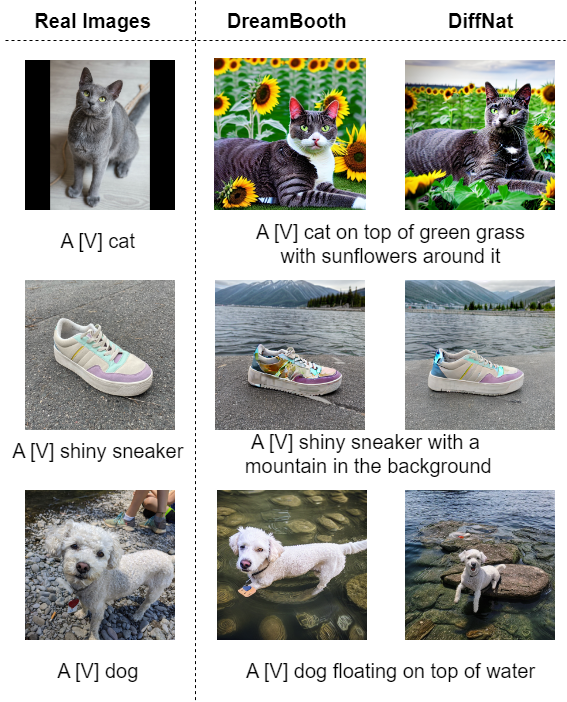}
    \caption{\textcolor{blue}{Comparison of DreamBooth and DiffNat. DiffNat generated images have better visual quality.}}
    \label{fig:diffnat_db_compare}
\end{figure*}

%\subsection{DiffNat generated images}
In Fig.~\ref{fig:diffnat_exp}, we visualize some of the DiffNat generated images using various text-prompts. The generated images capture the context of the text-prompt and also retain naturalness.
We have also provided qualitative comparison w.r.t Dreambooth in Fig.~\ref{fig:diffnat_db_compare}.

\begin{figure*}
    \centering
    \includegraphics[scale=0.45]{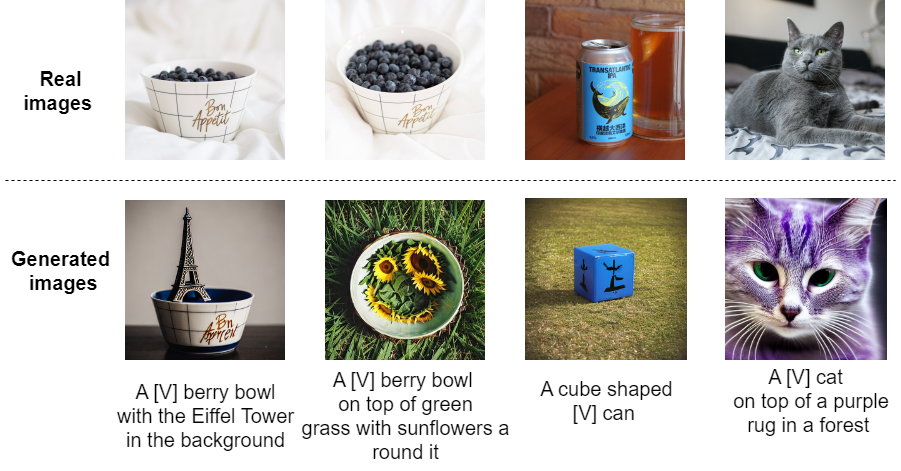}
    \caption{\textcolor{blue}{Failure cases of DiffNat. Instead of generating ``A [V] berry bowl with the Eiffel Tower in the background'', our method generates image with the Eiffel Tower in the berry bowl. Also, while generating ``A [V] cat on top of a purple rug in a forest'', it generates a purple [V] cat, which shows the color bias w.r.t the text-prompt of the model.}}
    \label{fig:failure_supple}
\end{figure*}

\section{Failure cases}
We also present some of the failure cases of DiffNat in Fig.~\ref{fig:failure_supple}. E.g., our model fails to generate images of ``A [V] berry bowl with the Eiffel Tower in the background'', but actually generates images with ``the Eiffel Tower'' in the berry bowl. Similarly, the model fails to generate ``A cube shaped [V] can'', since these object do not appear in the training set. The model also fails to generate ``A [V] cat on top of a purple rug in a forest'' and instead generated some version of purple cat.

% \subsection{Wavelet decomposition of images}

% In this section, we provide some visualization of wavelet decomposition components of natrual images in Fig.~\ref{fig:bag_wavelet}. As shown in Fig.~\ref{fig:bag_wavelet}, the LL and HH subbands capture the low-frequency and high-frequency component respectively. HL and LH subbands capture horizontal and vertical details respectively. 

% \begin{figure*}
%     \centering
%     \includegraphics[scale=0.45]{figures/color_mod_supple.png}
%     \caption{Color modification of object using text prompts.} 
%     \label{fig:color_modify}
% \end{figure*}

%%%%%%%%%%%%  added in the rebuttal %%%%%%%%%%%%%%%%%%
\textcolor{blue}{
\section{Computation complexity}
Here we analyze the computational complexity of the proposed KC loss. Suppose, given a batch of N images.
We need to perform DWT of each images using k different filters. Since, DWT for 'haar' wavelet can be done in linear time, the complexity of performing DWT with k filters can be done in $\mathcal{O}(Nk)$ time. Now, calculating the difference between maximum and minimum kurtosis can be done in linear time, therefore, the computational complexity of calculating KC loss is $\mathcal{O}(Nk)$. This minimal overhead of computing KC loss can be observed in the training time analysis provided next.
}

\textcolor{blue}{
\section{Training time analysis}
The run time analysis has been provided in Table.~\ref{table:training_time_analysis}. Note that the experiments for Dreambooth, Custom diffusion, DDPM have been performed on a single A5000 machine with 24GB GPU. We have performed guided diffusion (GD) and latent diffusion (LD) experiments on a server of 8 24GB A5000 GPUs. The experimental results in Table.~\ref{table:training_time_analysis} show that incorporating KC loss induces minimum training overhead.  
}

\begin{figure}
    \centering
    \begin{subfigure}{0.3\textwidth}
       \includegraphics[width=\linewidth]{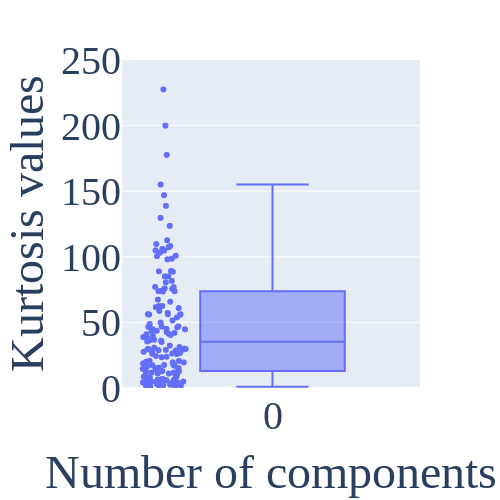}
        \caption{Average kurtosis of Dreambooth images}
    \end{subfigure}
    \begin{subfigure}{0.3\textwidth}
        \includegraphics[width=\linewidth]{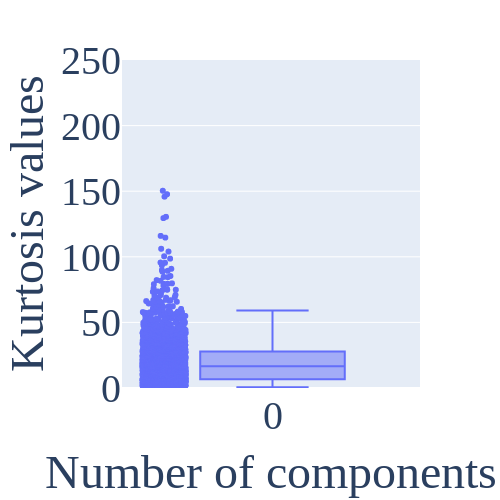}
        \caption{Avg. kurtosis (trained with Dreambooth + KC loss)}
    \end{subfigure}
    \begin{subfigure}{0.3\textwidth}
        \includegraphics[width=\linewidth]{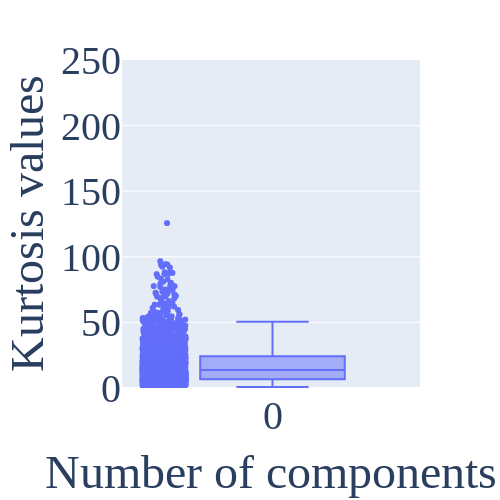}
        \caption{Average kurtosis of Natural images}
    \end{subfigure}
    \caption{\textcolor{blue}{Average kurtosis analysis of Dreambooth, DiffNat and natural images over the dataset used in Dreambooth. From this analysis, it is evident that Dreambooth generated images have higher kurtosis deviation. Integrating KC loss reduces the kurtosis deviation to preserve the naturalness of the generated images. Natural images have more concentrated kurtosis values.}}
    \vspace{-0.4cm}
    \label{fig:average_kurtosis}
\end{figure}

\begin{figure}
    \centering
    \begin{subfigure}{0.3\textwidth}
       \includegraphics[width=\linewidth]{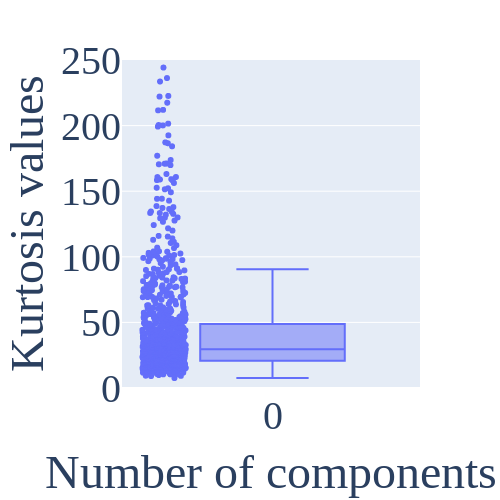}
        \caption{Average kurtosis of DDPM images}
    \end{subfigure}
    \begin{subfigure}{0.3\textwidth}
        \includegraphics[width=\linewidth]{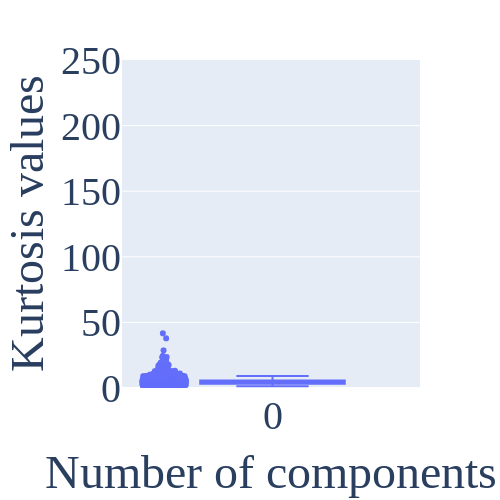}
        \caption{Average kurtosis of images trained with DDPM + KC loss}
    \end{subfigure}
    \begin{subfigure}{0.3\textwidth}
        \includegraphics[width=\linewidth]{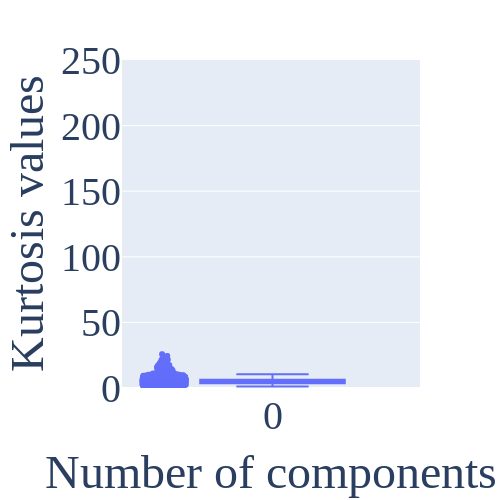}
        \caption{Average kurtosis of Natural images}
    \end{subfigure}
    \caption{\textcolor{blue}{Average kurtosis analysis of DDPM framework trained on Oxford flowers dataset. From this analysis, it is evident that DDPM generated images have higher kurtosis deviation. Integrating KC loss reduces the kurtosis deviation to preserve the naturalness of the generated images. Natural images have more concentrated kurtosis values.}}
    \vspace{-0.4cm}
    \label{fig:average_kurtosis_DDPM}
\end{figure}

\begin{figure}
    \centering
    \begin{subfigure}{0.3\textwidth}
       \includegraphics[width=\linewidth]{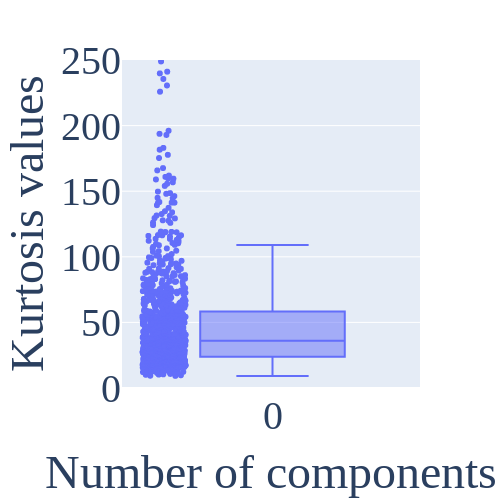}
        \caption{Average kurtosis of GD generated images}
    \end{subfigure}
    \begin{subfigure}{0.3\textwidth}
        \includegraphics[width=\linewidth]{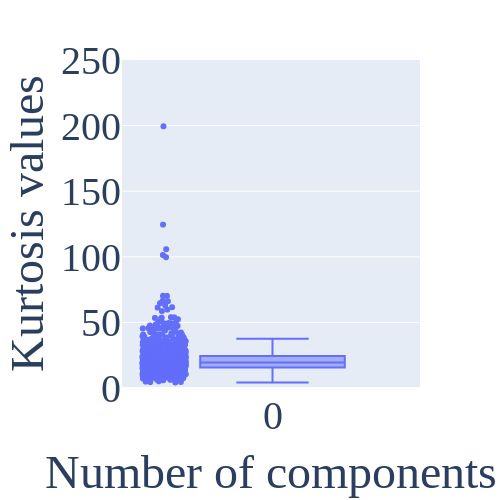}
        \caption{Average kurtosis of images trained with GD + KC loss}
    \end{subfigure}
    \begin{subfigure}{0.3\textwidth}
        \includegraphics[width=\linewidth]{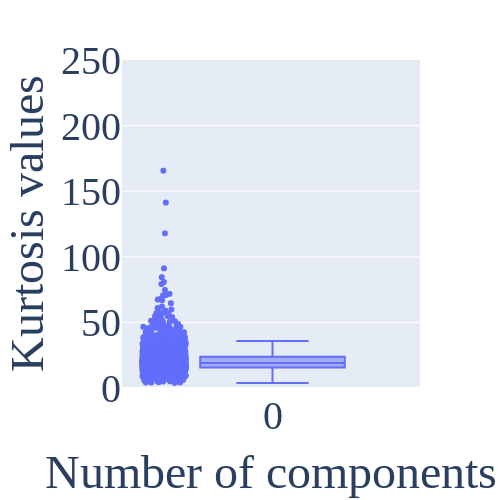}
        \caption{Average kurtosis of Natural images}
    \end{subfigure}
    \caption{\textcolor{blue}{Average kurtosis analysis of guided diffusion (GD) framework trained on FFHQ dataset. From this analysis, it is evident that GD generated images have higher kurtosis deviation. Integrating KC loss reduces the kurtosis deviation to preserve the naturalness of the generated images. Natural images have more concentrated kurtosis values.}}
    \vspace{-0.4cm}
    \label{fig:average_kurtosis_GD}
\end{figure}

\textcolor{blue}{
\section{Kurtosis analysis}
To verify the efficacy of the proposed KC loss, we perform average kurtosis analysis in this section.
we compute the average kurtosis deviation of DWT filtered version of images from the dataset and plot them in Fig.~\ref{fig:average_kurtosis}, Fig.~\ref{fig:average_kurtosis_DDPM} and Fig.~\ref{fig:average_kurtosis_GD}.
E.g., in case of dreambooth task, we compute the kurtosis statistics of bandpass filtered version of natural images from Dreambooth dataset, images generated by Dreambooth and images generated by DiffNat (i.e., adding KC loss) and plot it in Fig.~\ref{fig:average_kurtosis}. We observe that the Dreambooth generated images (Fig.~\ref{fig:average_kurtosis} (a)) have highest kurtosis deviation. The average deviation is least for natural images (Fig.~\ref{fig:average_kurtosis} (c)) and adding KC loss reduces the kurtosis deviation (Fig.~\ref{fig:average_kurtosis} (b)). Similar trends can be observed for DDPM (Fig.~\ref{fig:average_kurtosis_DDPM}), guided diffusion (Fig.~\ref{fig:average_kurtosis_GD}) as well. 
Adding KC loss improves image quality has been verified both qualitatively and quantitatively in the paper. 
This analysis verifies minimizing kurtosis loss improves diffusion image quality.}

\begin{figure*}
    \centering
    \includegraphics[scale=0.1]{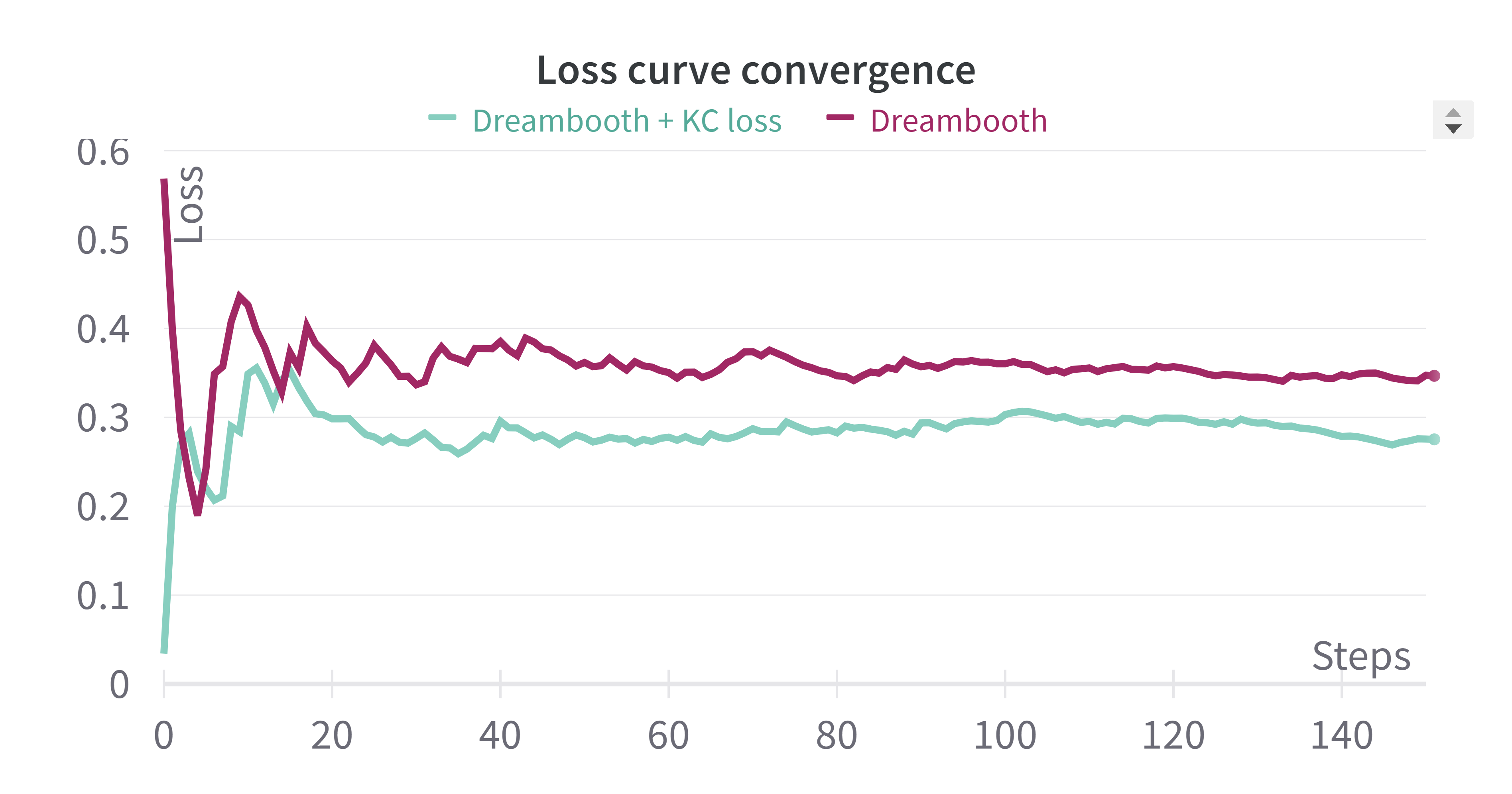}
    \caption{\textcolor{blue}{Loss curve convergence of Dreambooth.}}
    \label{fig:loss_curve_dreambooth}
\end{figure*}

\textcolor{blue}{
\section{Convergence analysis}
The main idea of the diffusion model is to train a UNet, which learns to denoise from a random noise to a specific image distribution. More denoising steps ensure a better denoised version of the image, e.g., DDPM~\cite{ho2020denoising}, LDM~\cite{karras2022elucidating}. In proposition 1 (main paper), we show that minimizing projection kurtosis further denoise input signals. Therefore, KC loss helps in the denoising process and improves the convergence speed. We have shown that adding KC loss improves the loss to converge faster for Dreambooth task in Fig.~\ref{fig:loss_curve_dreambooth}.
}

%\begin{wraptable}{r}{0.5\textwidth}
\begin{table}[!h]
  %\scriptsize
  \small
  \centering
  \vspace{-10pt}
  \caption{\textcolor{blue}{Training time analysis}}
  \scalebox{0.95}{
  \begin{tabular}{lcc}
  \toprule
  {\bf Method} & {\bf dataset} & {\bf Training time} \\ 
  \midrule
  {DreamBooth~\cite{ruiz2022dreambooth}} & {5-shot finetuning} & {10 min 21s}\\
  {DreamBooth~\cite{ruiz2022dreambooth} + KC loss} & {5-shot finetuning} & {11 min 30s}\\
  \hline
  {Custom Diffusion~\cite{kumari2022multi}} & {5-shot finetuning} & {6m 43s}\\  
  {Custom Diffusion~\cite{kumari2022multi} + KC loss} & {5-shot finetuning} & {7m 11s} \\
  \hline
  {DDPM~\cite{ho2020denoising}} & {CelebAfaces} & {2d 8h 21m}\\  
  {DDPM~\cite{ho2020denoising} + KC loss} & {CelebAfaces} & {2d 9h 19m} \\
  \hline
  {DDPM~\cite{ho2020denoising}} & {CelebAHQ} & {21h 48m}\\  
  {DDPM~\cite{ho2020denoising} + KC loss} & {CelebAHQ} & {22h 40m } \\
  \hline
  {DDPM~\cite{ho2020denoising}} & {Oxford flowers} & {6h 17m}\\  
  {DDPM~\cite{ho2020denoising} + KC loss} & {Oxford flowers} & {6h 39m} \\
  \hline
  {GD~\cite{dhariwal2021diffusion}} & {FFHQ} & {23h 10m}\\  
  {GD~\cite{dhariwal2021diffusion} + KC loss} & {FFHQ} & {1d 1h 29m} \\
  \hline
  {LD~\cite{karras2022elucidating}} & {FFHQ} & {20h 15m}\\  
  {LD~\cite{karras2022elucidating} + KC loss} & {FFHQ} & {22h 40m} \\
  \bottomrule
  \end{tabular}}
  \label{table:training_time_analysis}
  \vspace{-10pt}
 \end{table}
%\end{wraptable}

\begin{figure*}
    \centering
    \includegraphics[scale=0.7]{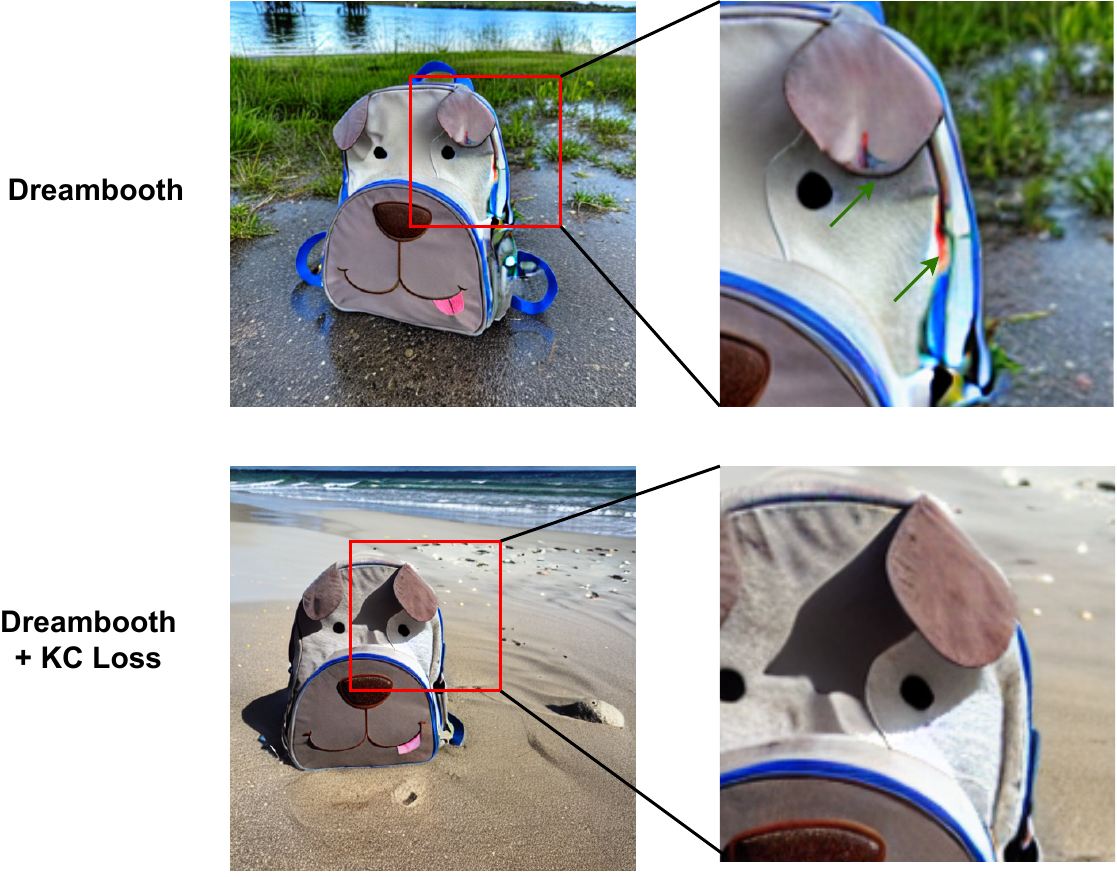}
    \caption{\textcolor{blue}{Qualitative comparison of with/without KC loss in Dreambooth. The bottom image (with KC loss) shows better image quality and shadows (best viewed in color).}}
    \label{fig:dogbag_db_diffnat}
\end{figure*}

\begin{figure*}
    \centering
    \includegraphics[scale=0.7]{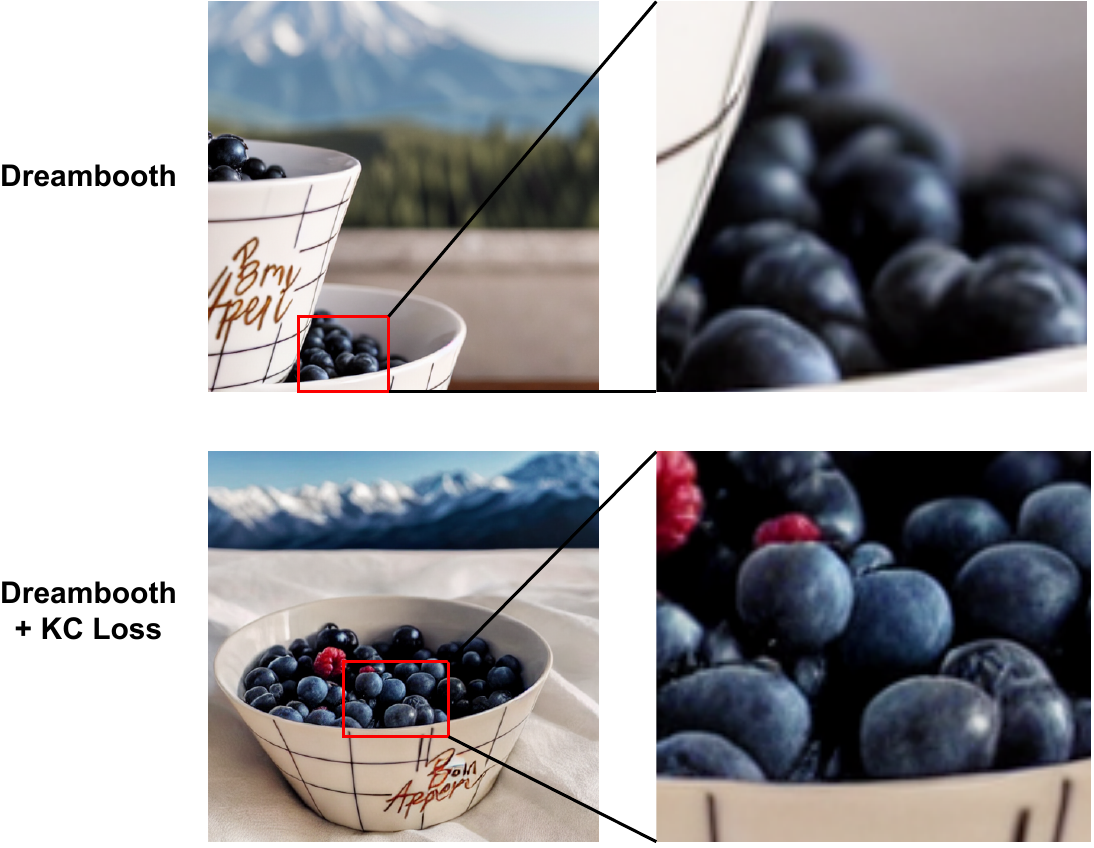}
    \caption{\textcolor{blue}{Qualitative comparison of with/without KC loss in Dreambooth. The bottom image (with KC loss) shows better image quality and reflections on the bowl full of berries (best viewed in color).}}
    \label{fig:berries_db_diffnat}
\end{figure*}

\begin{figure*}
    \centering
    \includegraphics[scale=0.75]{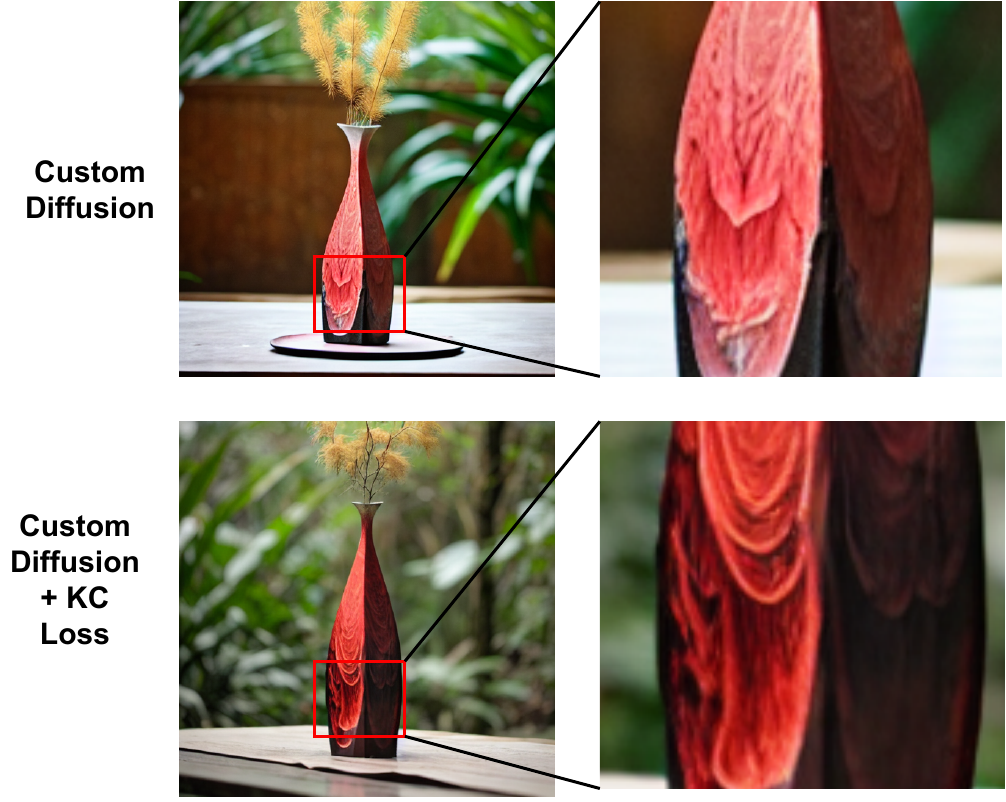}
    \caption{\textcolor{blue}{Qualitative comparison of with/without KC loss in Custom diffusion. The bottom image (with KC loss) shows better image quality in terms of color vividness and contrast (best viewed in color).}}
    \label{fig:vase_cd_diffnat}
\end{figure*}

\begin{figure*}
    \centering
    \includegraphics[scale=0.75]{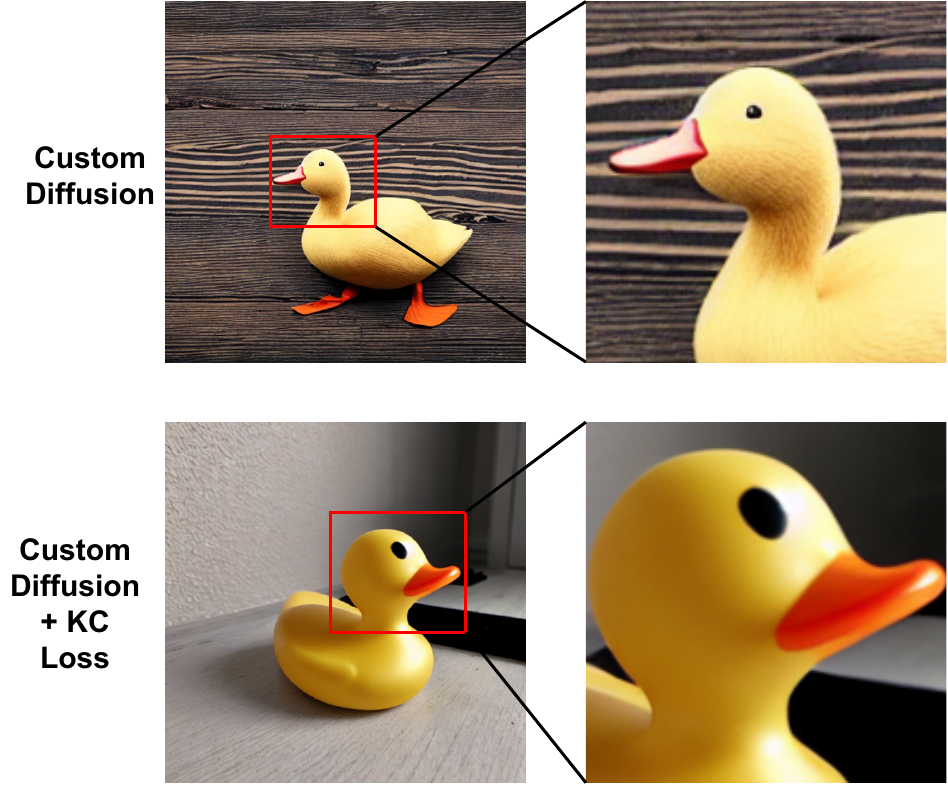}
    \caption{\textcolor{blue}{Qualitative comparison of with/without KC loss in Custom diffusion. The bottom image (with KC loss) shows better image quality in terms of detail and smoothness (best viewed in color).}}
    \label{fig:goose_cd_diffnat}
\end{figure*}

\begin{figure*}
    \centering
    \includegraphics[scale=0.7]{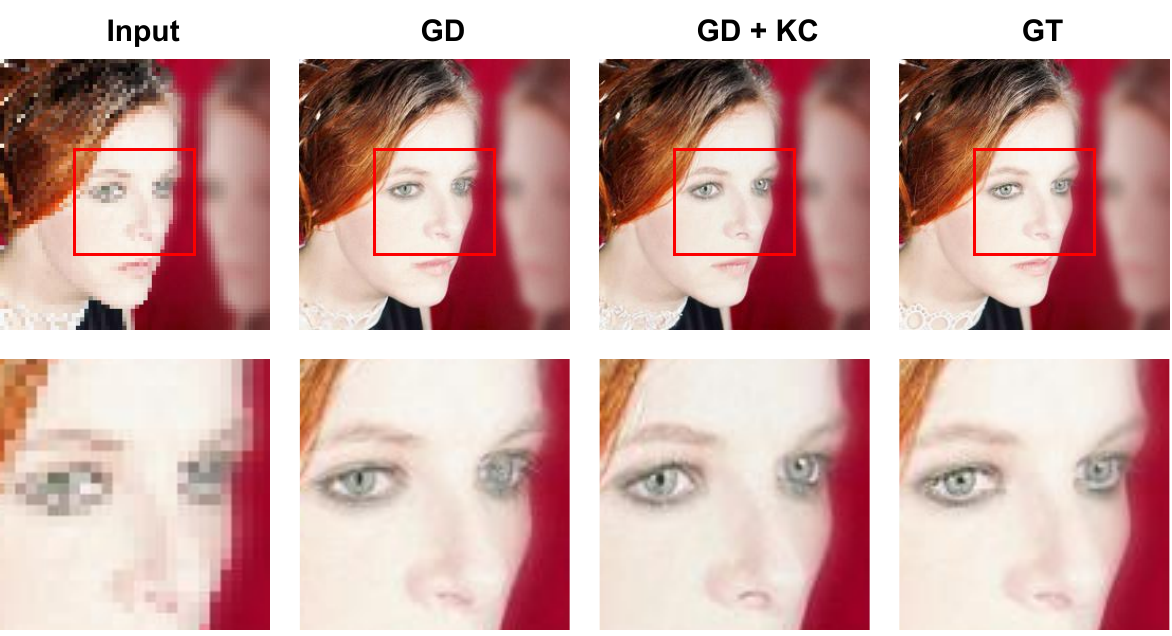}
    \caption{\textcolor{blue}{Qualitative comparison of with/without KC loss in guided diffusion (GD). The bottom image (with KC loss) has better eye and hair details (best viewed in color).}}
    %\caption{Zoomed view of using KC loss} 
    \label{fig:girl_gd_kc}
\end{figure*}

\begin{figure*}
    \centering
    \includegraphics[scale=0.5]{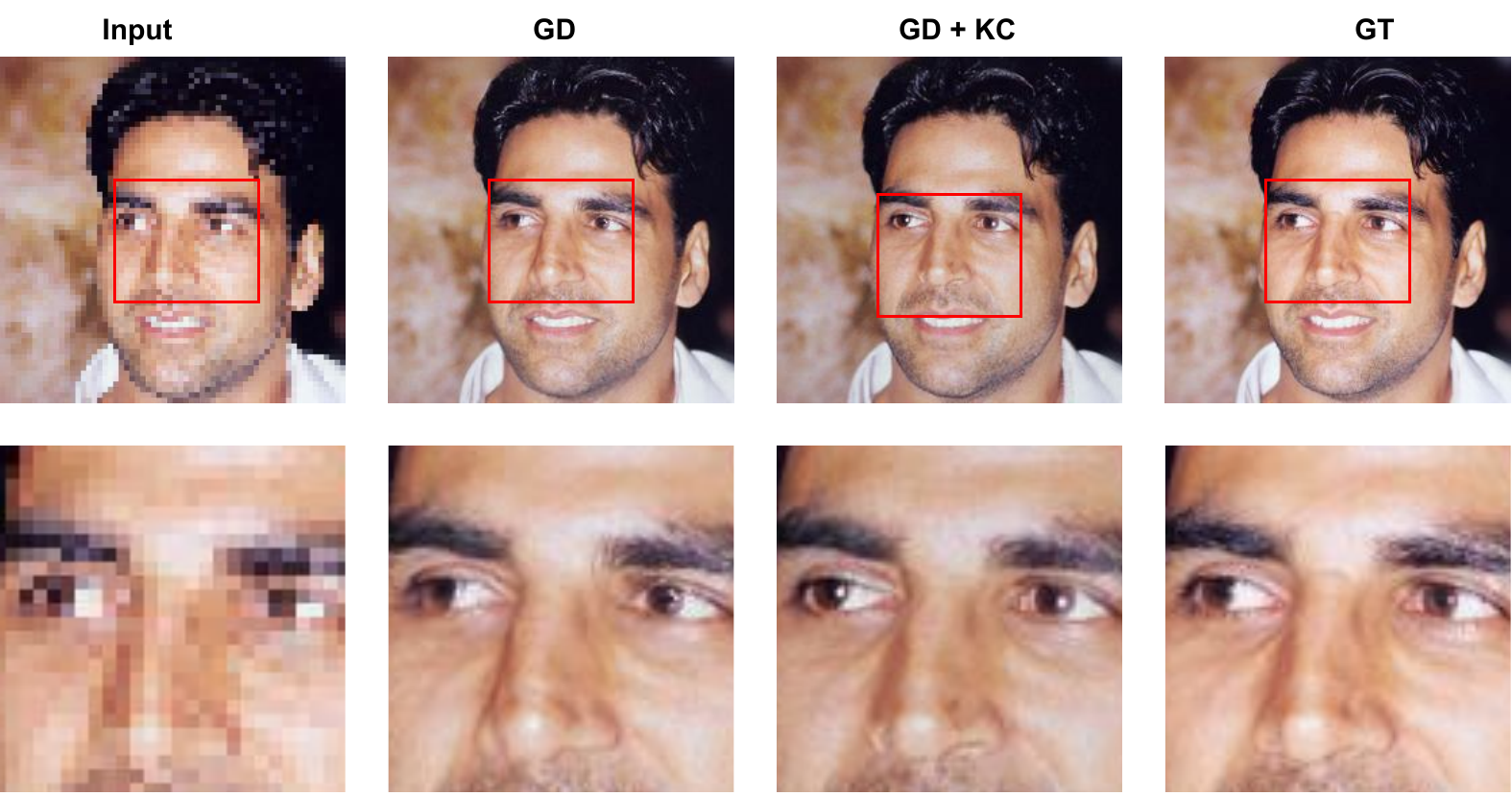}
    %\caption{Zoomed view of using KC loss} 
    \caption{\textcolor{blue}{Qualitative comparison of with/without KC loss in guided diffusion (GD). The bottom image (with KC loss) has better eye details and skin smoothness (best viewed in color).}}
    \label{fig:actor_gd_kc}
\end{figure*}

\begin{figure*}
    \centering
    \includegraphics[scale=0.6]{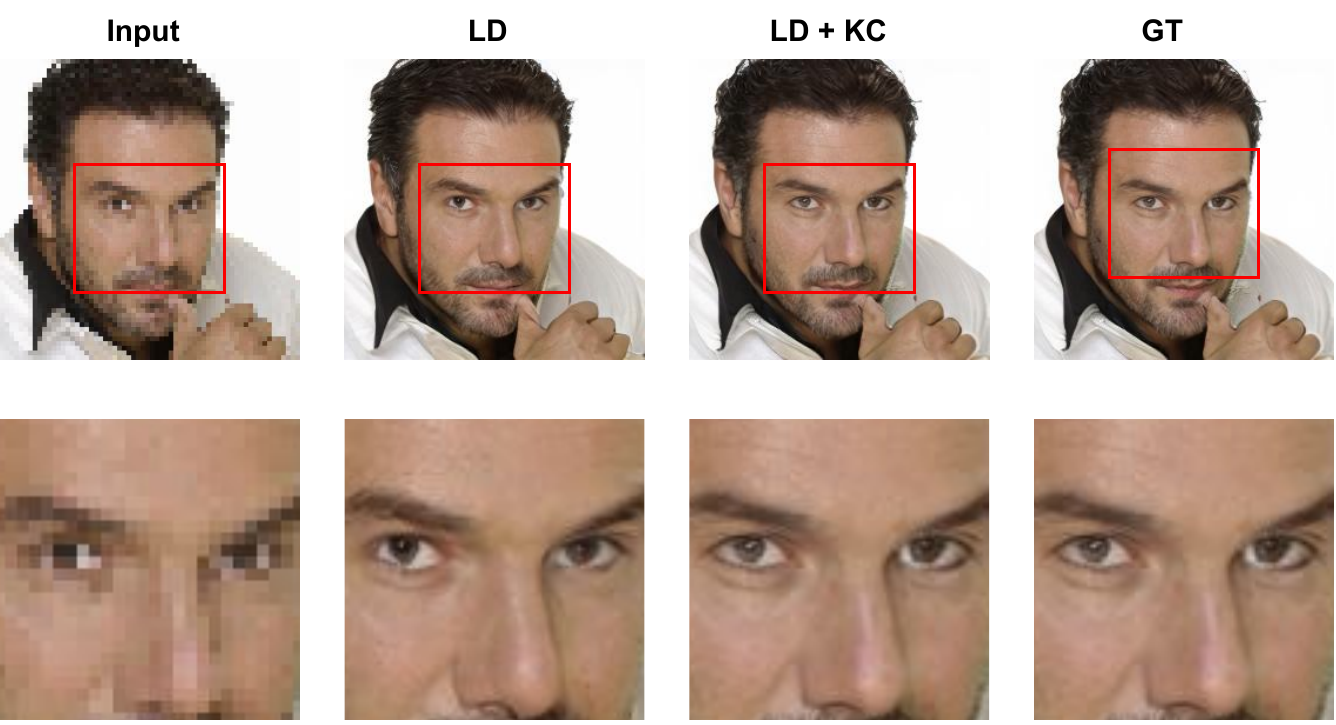}
    \caption{\textcolor{blue}{Qualitative comparison of with/without KC loss in Latent diffusion (LD). The bottom image (with KC loss) has higher similarity w.r.t the ground truth in terms of left eye and skin color (best viewed in color).}}
    \label{fig:guy_ld_kc}
\end{figure*}

\begin{figure*}
    \centering
    \includegraphics[scale=0.6]{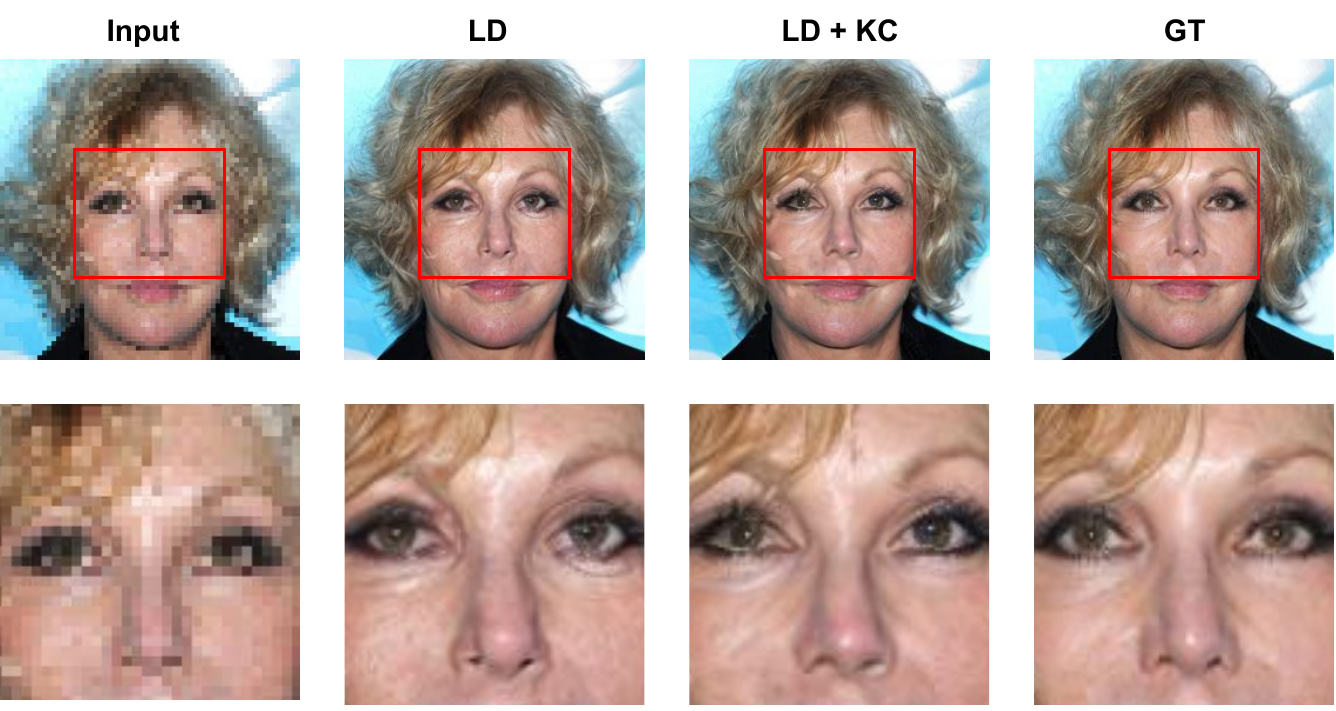}
    \caption{\textcolor{blue}{Qualitative comparison of with/without KC loss in Latent diffusion (LD). The bottom image (with KC loss) has better eye details and skin smoothness (best viewed in color).}}  
    \label{fig:aunty_ld_kc}
\end{figure*}

\textcolor{blue}{
\section{Qualitative analysis}
In this section, we provide more qualitative analysis to show that adding KC loss improves image quality.
Zoomed view of the generated images are shown to compare w.r.t the baselines in Fig.~\ref{fig:dogbag_db_diffnat}, Fig.~\ref{fig:berries_db_diffnat}, Fig.~\ref{fig:vase_cd_diffnat}, Fig.~\ref{fig:goose_cd_diffnat}, Fig.~\ref{fig:girl_gd_kc}, Fig.~\ref{fig:actor_gd_kc}, Fig.~\ref{fig:guy_ld_kc}, Fig.~\ref{fig:aunty_ld_kc}. Details are provided in the caption.
}

%%%%% additional results %%%

%%%%%%%%%%%%%%%%%%%%%%%%%%%%%%%%%%%%%%%%%%%%%%%%%%%%%%%%%%%%%%%%%%%%%%%%%%%

% \subsection{Property modification}

% We show results of generating different variations of objects w.r.t color in Fig.~\ref{fig:color_modify}. We can generate different color variants of ``[V] backpack'' object using the text-prompt only.

% \begin{figure}
%     \centering
%     \includegraphics[scale=0.4]{figures/wavelet_image.png}
%     \caption{Wavelet Transformed components of a natural image. LL and HH subband captures the low-frequency and high frequency details. Other subbands, e.g., LH, HL also captures vertical and horizontal details of the image.} 
%     \label{fig:bag_wavelet}
% \end{figure}

% \subsection{Comparison with SOTA}

% \textcolor{blue}{Here, we compare DreamBooth~\cite{ruiz2022dreambooth}, custom diffusion~\cite{kumari2022multi}, GD~\cite{dhariwal2021diffusion}, LDM~\cite{rombach2022high} and DiffNat generated images as shown in Fig.~\ref{fig:diffnat_db_compare}. We notice that DiffNat generated images are of better visual quality and have less distortions in the images.}

\end{document}